\PassOptionsToPackage{dvipsnames}{xcolor}
\documentclass[11pt,letterpaper]{mystyle}
\usepackage{fancyhdr}
\usepackage[utf8]{inputenc} 
\usepackage[T1]{fontenc}
\usepackage[numbers]{natbib}
\usepackage{adjustbox}
\usepackage{hyperref}
\usepackage[edges]{forest}
\usepackage{subcaption}
\usepackage{soul}
\usepackage{cleveref}
\usepackage{multirow}
\usepackage[utf8]{inputenc} 
\usepackage{booktabs}       
\usepackage{amsfonts}       
\usepackage{nicefrac}      
\usepackage{microtype}     
\usepackage[dvipsnames]{xcolor}
\usepackage{colortbl}
\usepackage{enumitem}
\usepackage{amsmath}  
\usepackage{amssymb} 
\usepackage{wrapfig}
\usepackage{courier}
\usepackage{twemojis}
\usepackage{bxcoloremoji}
\usepackage{CJKutf8}
\usepackage{tabularx}
\usepackage{longtable}
\usepackage{float}
\usepackage{rotating}
\usepackage{fontawesome5}
\usepackage{CJKutf8}
\usepackage{setspace}
\newcommand{\ghlink}[1]{\faIcon{github}\,\href{#1}{GitHub}}

\newcommand{\hflink}[1]{%
\includegraphics[height=1.em]{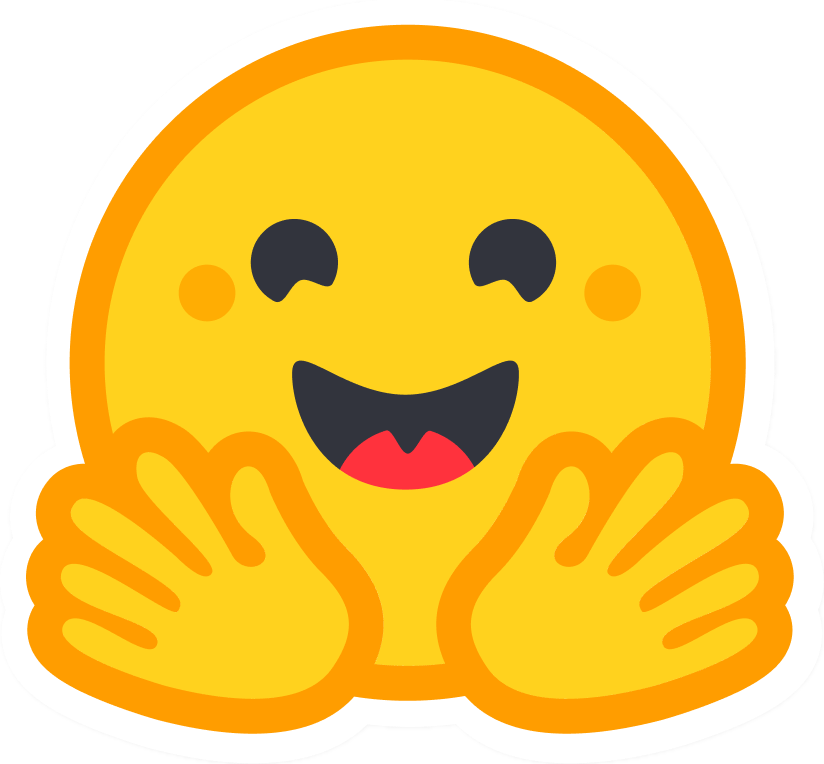}\,\href{#1}{HuggingFace}%
}

\newcommand{\weblink}[1]{\faIcon{globe}\,\href{#1}{Website}}

\usepackage{array}
\usepackage{graphicx}

\usepackage{booktabs} 
\usepackage{array} 
\usepackage{threeparttable}

\fancypagestyle{headstyle}{
    \fancyhead[L]{
    }

}

\crefformat{section}{\S#2#1#3} 
\crefformat{subsection}{\S#2#1#3}
\crefformat{subsubsection}{\S#2#1#3}

\definecolor{lightcoral}{rgb}{0.94, 0.5, 0.5}
\definecolor{darkpastelgreen}{rgb}{0.01, 0.75, 0.24}

\definecolor{hidden-red}{RGB}{205, 44, 36}
\definecolor{hidden-blue}{RGB}{194,232,247}
\definecolor{hidden-orange}{RGB}{243,202,120}
\definecolor{hidden-green}{RGB}{34,139,34}
\definecolor{hidden-pink}{RGB}{255,245,247}
\definecolor{hidden-black}{RGB}{20,68,106}
\definecolor{purple}{RGB}{144,153,196}
\definecolor{yellow}{RGB}{255,228,123}
\definecolor{hidden-yellow}{RGB}{255,248,203}
\definecolor{tkcolor}{RGB}{224,223,255}
\definecolor{darkblue}{rgb}{0, 0.40, 0.75}

\hypersetup{colorlinks=true, citecolor=darkblue, linkcolor=darkblue, urlcolor=darkblue}
\tcbset{
  aibox/.style={
    width=\linewidth,
    top=8pt,
    bottom=4pt,
    colback=blue!6!white,
    colframe=black,
    colbacktitle=black,
    enhanced,
    center,
    attach boxed title to top left={yshift=-0.1in,xshift=0.15in},
    boxed title style={boxrule=0pt,colframe=white,},
  }
}

\newtcolorbox{AIbox}[2][]{aibox,title=#2,#1}

\tcbset{
  takeawaybox/.style={
    width=\linewidth,
    top=8pt,
    bottom=4pt,
    colback=hidden-yellow,
    colframe=black,
    colbacktitle=black,
    enhanced,
    center,
    attach boxed title to top left={yshift=-0.1in,xshift=0.15in},
    boxed title style={boxrule=0pt,colframe=white,},
  }
}

\newtcolorbox{TakeawayBox}[2][]{takeawaybox,title=#2,#1}

\title{Modeling the Mental World for Embodied AI: A Comprehensive Review}

\author{
   \normalfont 
   Biyuan Liu$^{\textcolor{Maroon}{\alpha}\coloremojicode{2709}}$ \quad 
  Daigang Xu$^{\textcolor{Maroon}{\alpha}\coloremojicode{2709}}$ \quad 
   Lei Jiang$^{\textcolor{Maroon}{\alpha}}$ \quad 
   Wenjun Guo$^{\textcolor{Maroon}{\alpha}}$  \quad 
   Ping Chen$^{\textcolor{Maroon}{\alpha}}$ \\ \quad 
   
   \vspace{12pt}
   \small
   $^{\textcolor{Maroon}{\alpha}}$\textit{ZTE Corporation} \quad
   \\
   $^{\coloremojicode{2709}}$ \textit{Corresponding Author}
}

\begin{document}
\begin{CJK}{UTF8}{gbsn}
\begin{abstract}
  \textbf{\large Abstract:}
As the application of Embodied AI Agents in avatars, wearable devices, and robotic systems continues to deepen, their core research challenges have gradually shifted from physical environment interaction to the accurate understanding of human intention in social interactions. Traditional physical world models (PWM) focus on quantifiable physical attributes such as space and motion, failing to meet the needs of social intelligence modeling. In contrast, the Mental World Model (MWM), as a structured representation of humans’ internal mental states, has become the critical cognitive foundation for embodied agents to achieve natural human-machine collaboration and dynamic social adaptation. However, current MWM research faces significant bottlenecks: such as fragmented conceptual framework with vague boundaries between MWM and PWM, disparate element representation paradigms between psychological and computational neuroscience perspectives, disjointed reasoning mechanisms for the technical pathways and applicable scenarios of different Theory of Mind (ToM) reasoning paradigms, and detachment between evaluation and practice since existing benchmarks are mostly designed for static text scenarios and cannot meet the multimodal, real-time interaction needs of embodied agents.
To address these issues, this review systematically synthesizes over 100 authoritative studies to provide a comprehensive overview of MWM research for embodied AI. Its core contributions are threefold: First, it constructs a complete theoretical framework for MWM for the first time. Specifically, it distinguishes the essential differences between MWM and physical world models in state, observation, and action spaces. Second, it systematically defines the key components of MWM through Two paradigms for mental element representation (Strong psychological representations and Weak computational neuroscience representations). Third, it comprehensively analyzes two core ToM reasoning paradigms with 19 ToM methods, including ToM prompting and model-based inference paradigms. Finally, it also clarifies the integration trend of neuro-symbolic hybrid architectures, and synthesizes 26 ToM evaluation benchmarks (covering early static text, high-order text, and multimodal dynamic scenarios), providing researchers with references for technical selection and benchmark adoption. This work not only provides a unified cognitive framework for the theoretical research of social intelligence in embodied AI but also offers technical references for model design, benchmark selection, and ethical governance in engineering practice. Its aim is to promote the integration of embodied agents into human society and advance the in-depth development of human-machine collaborative interaction.

  \vspace{5mm}
  \vspace{4mm}

  \coloremojicode{1F4C5} \textbf{Date}: December 17th, 2025


  \coloremojicode{1F4E7} \textbf{Corresponding}: \href{mailto:byliu90@outlook.com}{byliu90@outlook.com}, \href{mailto:xu.daigang1@zte.com.cn}{xu.daigang1@zte.com.cn}\\
  \coloremojicode{1F4E7} \textbf{Main Contact}: \href{mailto:jiang.lei@zte.com.cn}{jiang.lei@zte.com.cn}, \href{mailto:guo.wenjun@zte.com.cn}{guo.wenjun@zte.com.cn}, \href{mailto:chen.ping8@zte.com.cn}{chen.ping8@zte.com.cn}

\end{abstract}
\maketitle

\vspace{3mm}
\pagestyle{headstyle}
\newpage
\tableofcontents

\newpage
\section{Introduction}

\begin{figure}[!b]
\centering
\includegraphics[width=1.0\textwidth]{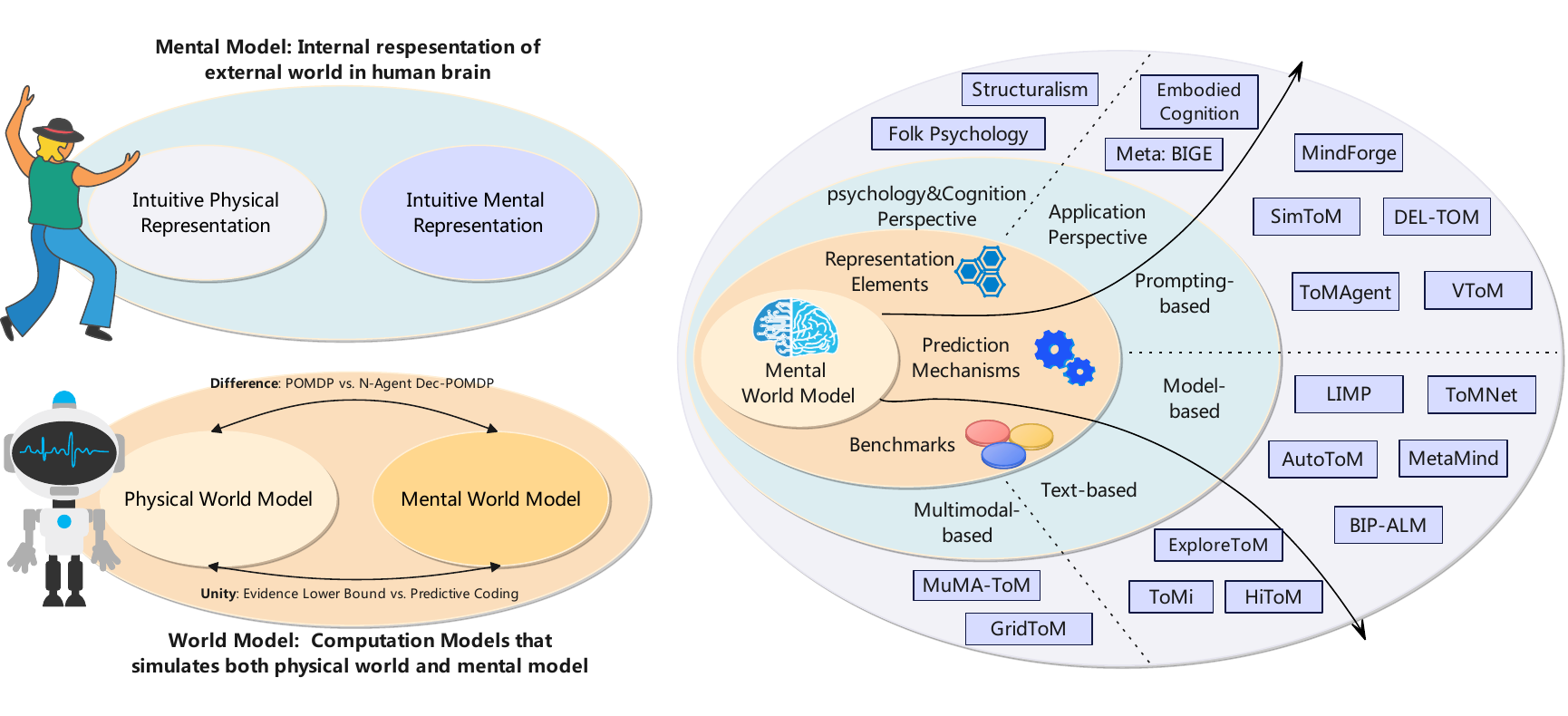}
\caption{Framework of Mental World Model for Embodied AI Agents. (a) Concept clarification of world model, physical world model, and mental world model. (b) Organization of this review.
}\label{fig_survey_framework}
\end{figure}

In recent years, Embodied AI Agents have achieved leaping progress in fields such as avatars (e.g., interactive characters in metaverse scenarios), wearable devices (e.g., the real-time environmental interaction function of Meta AI Glasses \cite{waisberg2024meta}), and service robots (e.g., home companion robots, industrial collaborative robots) \cite{duan2022survey}. The core research of early embodied AI agents focused on constructing physical world models \cite{ha2018world, ding2025understanding}, which involve perceiving the environment via sensors and planning motion trajectories to complete physical tasks (such as object grasping and path navigation). However, as application scenarios expand to human interaction, the core challenge for embodied AI agents has shifted from "handling the physical environment" to "understanding human social interaction". Specifically, they need to accurately infer humans' intentions, beliefs, and emotional states to achieve natural, adaptive social collaboration \cite{fung2025embodied}. For instance, a service robot must recognize the false belief that a user mistakenly placed a cup in the drawer to assist with retrieving the item correctly \cite{huang2021engaged}. Meta AI Glasses need to judge emotional tendencies through the user's tone of voice and facial expressions to provide tailored information recommendations \cite{waisberg2024meta}. This shift highlights the limitations of traditional PWMs, and there is an urgent need for a cognitive model oriented toward "human mentality" to serve as the foundation for social intelligence.

To address this need, it is first necessary to clarify the connotations and boundaries of core concepts. According to the definition in \cite{ding2025understanding}, a world model is a computational model that enables agents to simulate and predict the environment, with core functions of "\textit{constructing internal representations to understand world mechanisms}" and "\textit{predicting future states to guide decision-making}". It can be divided into two main branches: physical world models (PWMs) and mental world models (MWMs). Among these, the PWM focuses on objective physical laws: its state space only includes physical attributes such as position and material, its observation space consists of environmental physical signals, and it supports physical behaviors like object grasping and path planning \cite{cen2025worldvla}. Meanwhile, the MWM, as a key subset of the World Model, was formally defined by Fung et al. \cite{fung2025embodied} as a structured representation of "humans' internal mental states," covering subjective elements such as beliefs (including true/false beliefs), goals, intentions, and emotions, with the core function of simulating how humans understand the world and others.
The two models are closely interconnected. The PWM acts as a "real‑world anchor" for the MWM—human mental states, such as the belief that "the cup is on the table," are formed and updated through perception of the physical environment. Conversely, the MWM serves as a "social extension" of the PWM, enabling embodied AI agents to interpret the social meaning behind physical behaviors. For instance, the action of "picking up a cup" may correspond to the goal of "drinking water" or the intention of "handing it to someone else." Theoretically, the MWM stems from the theory of mental representation, which posits that cognitive agents simulate the external world through internal representations. The essence of social intelligence, in turn, lies in the capacity for "representing the representations of others."

Special clarification is needed regarding MWM's distinction from the mental model. The mental model is a cognitive psychology concept \cite{schwamb1990mental}, referring to the hybrid representation of external reality in the human brain (integrating physical perception and psychological inference). The impression of the surrounding world in our minds is merely a model—no one can visualize an entire world, government, or country in their head. instead, one selects certain concepts and the relationships between them, using these to represent the actual system \cite{craik1967nature, forrester1971counterintuitive}. For example, the Intuitive Physics Engine is a typical mental model that humans use to simulate the temporal evolution of the physical world \cite{battagliaintuitive}. In contrast, the MWM is an engineered model for embodied AI agents, specifically designed to model "human mental states" rather than a simple extension of the mental model. The connection between the two lies in this: the MWM draws on the core logic of "psychological inference" from the human mental model, but achieves machine computability through structured designs (such as symbolic beliefs and probability distributions).
The relationships between these concepts are illustrated in Figure \ref{fig_survey_framework}(a). However, existing research on world models has significant limitations: it either focuses on implicit representations of the external physical world (e.g., JEPA \cite{lecun2022path}, V-JEPA \cite{bardes2024revisiting}, V-JEPA2 \cite{assran2025v}), or relies on generative models to predict future states of the physical world (e.g., CityGPT \cite{feng2025citygpt}, Genie \cite{bruce2024genie}, DreamGen \cite{jang2025dreamgen}), neither involves the modeling and deduction of "mental laws."

Although the concept of the MWM has been clarified, current research still faces three core gaps: Firstly, fragmented theoretical frameworks—existing studies mostly discuss mental elements in a scattered manner (e.g., focusing solely on beliefs), lacking systematic modeling of the coupled relationships between "beliefs-emotions-goals" and failing to establish a unified mathematical framework integrating physical and mental world models. Secondly, lack of comparison among reasoning paradigms—the two major paradigms of Theory of Mind (ToM) reasoning (Prompting and Model-Based) each have their own advantages and disadvantages, yet existing works have not systematically analyzed their technical characteristics, applicable scenarios, or integration pathways. Third, evaluation benchmarks are disconnected from embodied needs—early benchmarks (e.g., ToMi \cite{le2024revisiting}) are limited to static text, while multimodal benchmarks (e.g., MuMA-ToM \cite{shi2025muma}) have made certain breakthroughs but suffer from insufficient data authenticity and a lack of online interactive evaluation, making it difficult to reflect the actual needs of embodied agents for reasoning while interacting. Furthermore, existing reviews \cite{saritacs2025systematic, marchetti2025artificial} mostly focus on evaluating the ToM capabilities of large language models (LLMs) and fail to integrate the three core dimensions of "representation-reasoning-benchmark" from the cognitive perspective of embodied agents.

To address the aforementioned gaps, this paper conducts a systematic review centered on how MWM supports the social intelligence of embodied AI agents. The overall structure is illustrated in Figure \ref{fig_survey_framework}: First, organizing the theoretical framework—by synthesizing over 100 authoritative studies, it clarifies the element representation paradigms of the MWM (strong psychological representations and weak computational neuroscience representations), and demonstrates the differences and unity between MWM and the PWM by leveraging Partially Observable Markov Decision Processes (POMDP) \cite{smallwood1973optimal, curtis2025llm} and Predictive Coding Theory \cite{millidge2021predictive}. Second, comparing reasoning paradigms—it conducts an in-depth analysis of the typical methods, trade-offs (interpretability vs. efficiency), and neuro-symbolic integration trends of the two core paradigms: the ToM Prompting paradigm (which stimulates the implicit capabilities of large language models (LLMs)) and the Model-Based paradigm (which explicitly constructs symbolic models). Third, tracing the evolution of benchmarks—following the logic of "static text $\rightarrow$ high-order text $\rightarrow$ multimodal dynamic," it summarizes the advantages and limitations of 25 benchmarks and identifies the core reasons for their insufficient adaptability to embodied scenarios. The key difference between this paper and existing reviews lies in its realization of full-chain integration of "theory-method-benchmark" based on the cognitive needs of embodied AI agents, rather than focusing solely on single-dimensional capability evaluation.
The main contributions of this paper can be summarized as follows:
\begin{itemize}
    \item  \textbf{Establishing a unified theoretical framework}. Addressing the fragmentation of the MWM theory, this paper systematically clarifies the essential differences (the state space includes psychological attributes, and the observation space includes introspective signals) and connections (the physical world model serves as a "real-world anchor," while the MWM acts as a "social extension") between the MWM and the PWM. It further achieves the mathematical unification of the two via Predictive Coding \cite{spratling2017review, millidge2021predictive}, filling the theoretical gap in the field.
    \item \textbf{Providing technical selection references}. Following the evolutionary logic of “static text$\rightarrow$high-order text$\rightarrow$multimodal dynamic interaction”,   this paper organizes 26 ToM benchmarks. By comparing 19 typical methods, it analyzes the technical characteristics of the Prompting and Model-Based paradigms. It also identifies two core integration pathways: "neural generation + symbolic verification" and "symbolic guidance + neural fine-tuning."
    \item \textbf{Identifying practicalization directions}. Focusing on key technical bottlenecks such as the dynamic update of mental states, multimodal information alignment, and the robustness of high-order reasoning, this paper also discusses ethical risks including excessive reliance on anthropomorphism, potential privacy leakage from multimodal signals, and model biases caused by pre-trained data. Several actionable future directions are proposed, providing guidance for the practical research on the social intelligence of embodied AI agents.
\end{itemize}

The remaining structure of this paper is organized as follows: Chapter 2 clarifies the differences and unity between the PWM)and the MWM from the perspectives of conceptual and mathematical connotations. Chapter 3 theoretically examines the element representation paradigms of the MWM (such as symbolic beliefs and probabilistic beliefs). Chapter 4 compares the typical methods and integration trends of the two major reasoning paradigms. Chapter 5 traces the evolutionary context of ToM evaluation benchmark, and analyzes their adaptability limitations. Chapter 6 summarizes the core technical challenges and proposes future research directions.


\section{Analysis of Difference and Unity between PWM and MWM}

This section aims to answer the following questions by analyzing the differences and unity between the PWM and the MWM:
\begin{tcolorbox}[colframe=black!70, colback=yellow!5, boxrule=1pt, arc=4mm]
\begin{itemize}
    \item Why is the human brain's world model a unified mental model, yet when converted into computational models, the physical and mental world models need to be studied separately? What are the differences between the two in terms of conceptual connotations and mathematical definitions?
    \item Do the physical and mental world models have unity, and can they be described by a unified mathematical model?
\end{itemize}
\end{tcolorbox}

\subsection{Difference Analysis}\label{section_physical_vs_mental}

\begin{table}[!htb]
    \centering
    \setlength{\belowcaptionskip}{0.2cm}
    \caption{Comparison between Physical World Model and Mental (Social \cite{zhou2025social,zhangsocial}) World Model.}
    \scalebox{0.75}{
    \begin{tabular}{p{4em}p{5cm}p{4cm}p{3cm}p{4.cm}p{3cm}}
    \toprule
   	World Models &State Space	& Observation Space&Action Space&  Supported Behaviors & Theoretical Foundations\\
   \midrule
Physical & Objective states of the physical world (position, size, material, etc.)& Physical observations of the environment &Physical actions&Physical interactions& Traditional World Models \cite{ha2018world, xiang2023language}, Reinforcement Learning \cite{sutton1998reinforcement}\\
Mental (Social) & Physical attributes + psychological attributes (beliefs, intentions, emotions, etc.)&External observations + introspective observations + memory& Physical actions + cognitive actions&Complex social behaviors (empathy, deception, norm enforcement, etc.)&ToM in cognitive science\cite{premack1978does,lakoff2024metaphors}\\

\bottomrule
    \end{tabular}}

    \label{table_physical_vs_mental_wm}
\end{table}

There is an essential difference between the PWM and the MWM (as well as Social World Model). As shown in Table \ref{table_physical_vs_mental_wm}, the traditional PWM \cite{ha2018world, xiang2023language} mainly focuses on the physical states of the environment and their transition rules, with its state space including observable physical attributes such as an object’s position, temperature, size, and material. Agents form action decisions by perceiving these physical information. In contrast, while retaining the ability to model physical states, the MWM further extends to modeling agents’ mental states, representing a qualitative leap. Specifically, the MWM not only needs to track "what the world is like" but also infer "what others think the world is like" and "what others want"—this involves the representation of other agents’ internal psychological states such as beliefs, intentions, emotions, desires, and moral values. To support such complex modeling, the MWM expands the agent’s observation space into two dimensions: external observations (information from the environment and other agents, such as dialogue content or behaviors) and introspective observations (the agent’s own internal states, such as goals and emotions), and incorporates memory retrieval into the cognitive action space. This design enables the MWM to capture the core capability in human social interactions—Theory of Mind (ToM), i.e., the ability to recursively reason about others’ psychological states—thereby supporting more realistic and intelligent social decisions, including complex social behaviors such as empathy, deception, forgiveness, and norm enforcement.

According to the definition proposed by Richens et al. from Google DeepMind \cite{richens2025general}, any agent capable of completing multi-step tasks (more than two steps) implicitly incorporates an environmental transition function that predicts environmental changes—this function is referred to as the agent’s world model. The PWM and MWM exhibit significant differences in the formal definition of the transition function. For the traditional PWM \cite{ha2018world, sakagami2023robotic}, it can be described using the standard POMDP \cite{smallwood1973optimal, curtis2025llm} framework as a five-tuple $(S_t, A_t, O_t, T, \Omega)$ at time $t$, where the state space $S_t$only includes physical attributes, the action space $A_t$d escribes physical actions, the observation space is $O_t$，and the observation function $\Omega$ and environmental transition function $T$ are expressed as follows. 
\begin{align}
\Omega^{ex}&: p(O_{t}^{ex}|A_{t}^{ex},S_t) \label{eq_as_to_o}\\ 
    T&: p(S_{t+1}|A_{t}^{phy},S_{t})\label{eq_physical_T}
\end{align}
Among them, the observation function \(\Omega\) maps the physical state \(S_t\) to external observations \(O_{t}^{ex}\) via external observation actions \(A_{t}^{ex}\), and the transition function $T$ models the evolution of the physical state from \(S_{t}\) to \(S_{t+1}\) as influenced by physical actions \(A_{t}^{phy}\). The world model can be described as an approximation of the environmental transition function.
\begin{equation}
    \hat{T} \approx T = P(S_{t+1} | A_t, S_t )
\end{equation}
The upper bound of this transition function is 
\begin{equation}
    |\hat{T} - T| \leq \sqrt\frac{2 \hat{T}(1-\hat{T})}{(n-1)(1-\delta)}
\end{equation}
where $n$ denotes the number of task steps and $\delta$ represents the average failure rate per step. In the case where $\delta\rightarrow0$ and $1\ll n$, the lower bound can be further approximated as $o(\delta/\sqrt{n})+o(1/n)$.As can be seen from Equation \eqref{eq_physical_T}, the transition of world states in a purely physical world model is only affected by physical actions $A_{t}^{phy}$ and historical world states $S_t$. Here, physical actions triggered by both living organisms and non-living entities are collectively referred to as physical actions, and external observation actions $A_{t}^{ex}$ of living organisms do not induce changes in world states.However, the physical world model lacks a description of the process by which observations of living organisms influence their physical actions (here we assume that living organisms possess mentality, and their actions are driven by mentality rather than reflexive phenomena \cite{domjan1993domjan}). Therefore, we need to model the mental world model to provide a mechanistic explanation for changes in world states caused by agents.

\begin{figure}[!htb]
\centering
\includegraphics[width=1\textwidth]{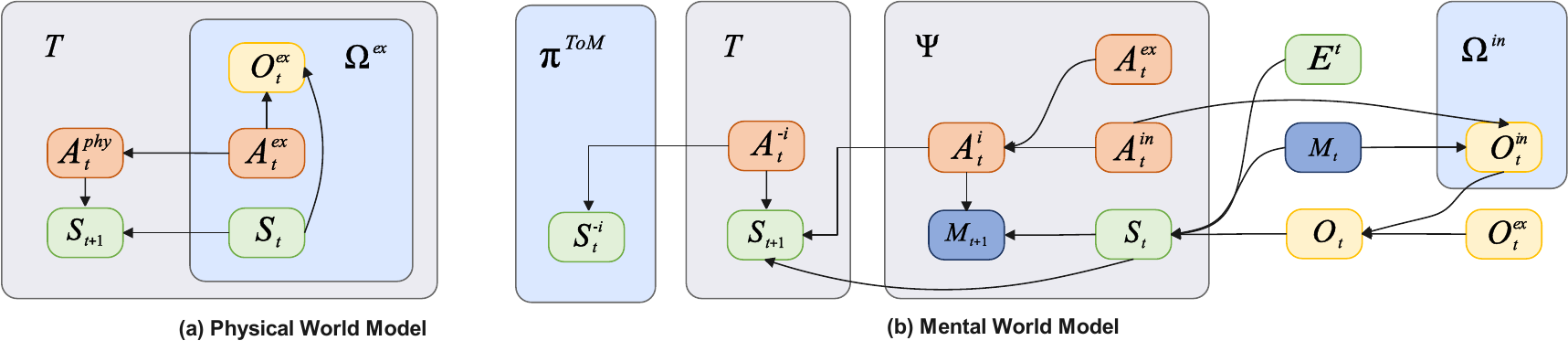}
\caption{Probabilistic Dependency Representation of State Variables in Physical and Mental World Models}\label{physical_vs_mental_wm}
\end{figure}

The observation space of each agent can be decomposed into \(O_t = \{O_t^{ex}, O_t^{in}\}\), where \(O_t^{ex}\) captures external physical and social information (e.g., "what someone said"), and \(O_t^{in}\) represents introspective observations—i.e., observations of mental states such as beliefs, goals, emotions, and moral preferences. In psychology and cognitive science, memory retrieval is regarded as a cognitive action \cite{weger2018introspection}, and introspective observations shape the process by which agents make decisions based on their own mental states \cite{wilson1991thinking, schwitzgebel2010introspection}, endowing agents with the ability to act reflectively rather than reactively.The state space needs to be decomposed into \(S_t = (E_t, M_t, O_t)\), where \(E_t\) denotes the physical state of the environment (in the PWM, \(s_t=E_t\)) and \(M_t\) represents the agent’s memory state.Since introspective observations rely on memory, a memory function is introduced—i.e., the process by which world states and actions collectively shape memory.

\begin{equation}
    \Psi: p(M_{t+1}|A_t,S_t) \label{eq_as_to_m}
\end{equation}
It should be noted that Equation \eqref{eq_as_to_m} and Equation \eqref{eq_as_to_o} have similar expressions, but observation and memory are clearly not equivalent.
Further, introspective observation can be described as
\begin{equation}
    \Omega^{in}: P(O_t^{in}|M_t,A_t^{in})
\end{equation}
where \(A_t^{in}\) denotes the action of introspective observation, and the state transition function of agent $i$ is defined as
\begin{equation}
    T: P(S_{t+1}^i|S_t^i,A_t^i,A_t^{-i})\label{single_world_T}
\end{equation}
where \(A_t^{-i}\) denotes the actions of other agents within the observation range of agent $i$. The above equation indicates that the state transition of an individual’s MWM now depends on the environmental state, external observations, introspective observations, memory, and the currently observed actions of other agents within the agent’s observation range, rather than merely physical laws. This formal difference essentially reflects a paradigm shift from "modeling how the world works" to "modeling how agents think about the world and each other." When the MWM is extended to a social world model, it needs to be further expanded to the N-agent Dec-POMDP framework \cite{bernstein2002complexity, nair2003taming} to characterize the mental interactions among multiple agents. First, the joint action space of $N$ agents is defined as \(A_t = \{A_t^1, ..., A_t^N\}\), the joint observation space is defined as \(O_t = \{O_t^1, ..., O_t^N\}\), and the state space is similarly decomposed into \(S_t = (E_t, M_t, O_t)\), where \(M_t = \{M_t^1, ..., M_t^N\}\) represents the memory states of all agents.For the i-th agent, treating memory retrieval as a cognitive action transforms the agent’s action policy into
\begin{equation}
    \pi_i: p(A_t^i|S_t^i)=p(A_t^i|E_t, M_t^i, O_t^i)
\end{equation}
That is, the action of agent $i$ at time $t$ depends on memory and observations. By analogy, the actions of the remaining agents \(p(A^{-i}_t | S_t^{-i})\) (i.e., predicting the actions of other agents via ToM) can be derived. The inference of mental states (i.e., Bayesian ToM \cite{baker2017rational}) can be described as
\begin{equation}
    \pi^{ToM}: p(S_t^{-i}|A^{-i}_t) \propto p(A^{-i}_t | S_t^{-i}) 
\end{equation}
Then the transition function of the social world model can be expressed as
\begin{equation}
    T:p(S_{t+1} | S_t, A^{-i}_t, A^i_t)\label{social_world_T}
\end{equation}
The above equation indicates that the state transition of the group-level MWM (i.e., the social world model) now depends on the mental state evolution of all agents. When comparing Equation \eqref{single_world_T} and Equation \eqref{social_world_T}, they are formally very similar. However, the former characterizes the mental evolution of an individual, while the latter describes the mental evolution of a group. Meanwhile, constrained by the limits of their observation space, individuals obtain most of their observations of other agents' actions through direct observation, whereas the social world model possesses more information and predicts agents' actions by virtue of ToM.

\subsection{Unity Analysis}

\begin{figure}[!htb]
\centering
\includegraphics[width=1\textwidth]{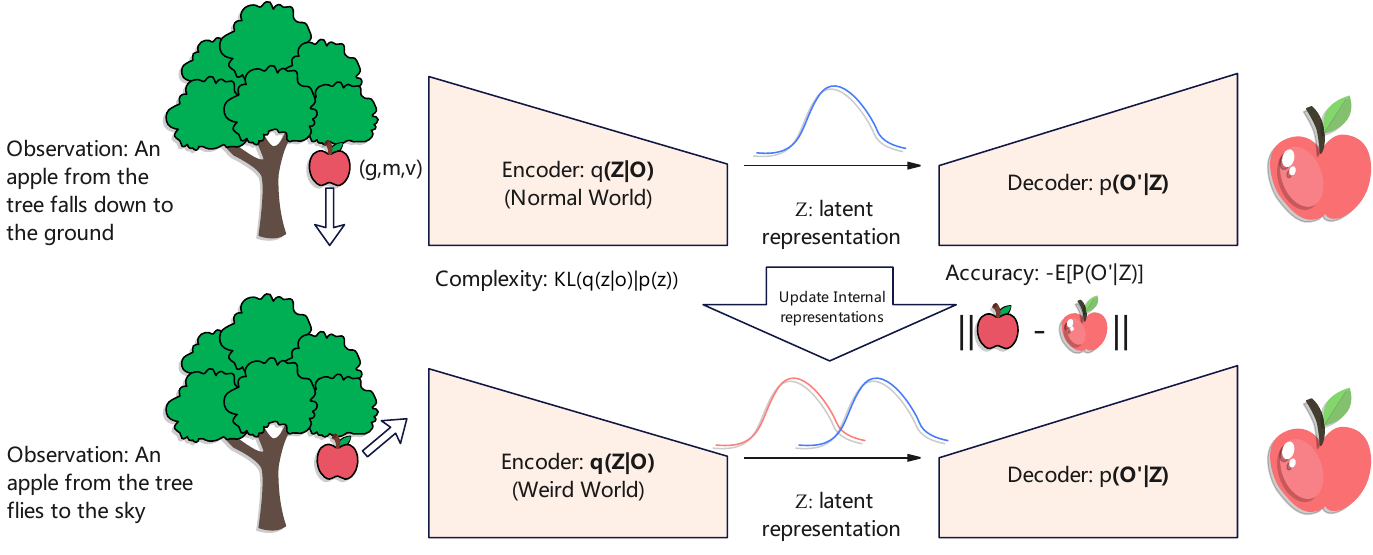}
\caption{An external observation $o$ is first internally represented as \(q(z|o)\) via a generative representation model, and then explicitly or implicitly decoded as \(p(o'|z)\). The discrepancy between the decoded output \(o'\) and the real observation $o$ is defined as the accuracy error. The second observation in the figure exaggeratedly illustrates the need to measure the complexity between the prior and the observed representation.$KL(q(z|o)|p(z))$. }\label{fig_pc_in_world_model}
\end{figure}

This subsection demonstrates the mathematical unity of the PWM and the MWM from the perspectives of generation and prediction by leveraging predictive coding and generative representation models. As shown in Figure \ref{fig_pc_in_world_model}, suppose an individual observes an event where an apple falls from a tree. With the help of the intuitive physics engine \cite{battagliaintuitive}, humans can roughly estimate the apple’s acceleration $g$ and mass $m$ by observing changes in the apple’s falling velocity $v$. This process converts observations of world states into belief distributions, thereby supporting humans’ physical prediction capabilities. Predictive coding \cite{millidge2021predictive} is a hierarchical cognitive framework based on Bayesian inference. Its essence lies in the following mechanism. An agent constructs a "generative model" of the environment and outputs predictive signals in a top-down manner. It simultaneously iteratively revises the model using bottom-up "prediction errors". Ultimately, it minimizes the "free energy" (or the sum of prediction errors) to achieve accurate modeling of the environment and efficient interaction with it. By means of predictive coding and the generative representation process illustrated in Figure \ref{fig_pc_in_world_model}, we can explain how the human brain fine-tunes internal representations to perform world prediction.

When observing an event of an apple falling in a Normal World, the Encoder performs implicit encoding on the observation $o$ to obtain a latent representation $z$, a process referred to as the inference process \(q(z|o)\). In the context of predictive coding, this is interpreted as the process of inferring the posterior distribution of hidden causes based on the observation o. Since the posterior distribution \(p(z|o)\) is difficult to compute directly, the inference process is formulated as \(q(z|o)\) rather than \(p(z|o)\). In turn, resampling $z$ from the learned implicit representation distribution and generating an output \(o'\) corresponding to the real observation via the Decoder constitutes the generative process \(p(o|z)\). In classical generative autoencoder models (e.g., VAE \cite{pinheiro2021variational}), the objective is to enable the model to "learn to generate samples consistent with real data", which is achieved by maximizing the Evidence Lower Bound (ELBO), namely:
\begin{equation}
    \text{ELBO}(\phi, \theta; o) = \underbrace{\mathbb{E}_{q_\phi(z|o)}\bigl[\log p_\theta(o|z)\bigr]}_{\text{Reconstruction Loss}} - \underbrace{\text{KL}\bigl[q_\phi(z|o) \mid\mid p(z)\bigr]}_{\text{Regularization Loss}}
\end{equation}
where \(\phi\) is used to parameterize the Encoder and \(\theta\) is used to parameterize the Decoder. We now interpret this process from the perspective of predictive coding. In predictive coding, the Free Energy Principle (FEP) \cite{friston2010free, millidge2021predictive} is leveraged to characterize the discrepancy between the brain’s actual observations and the predictions derived from its internal representations, namely:
\begin{equation}
    \text{FE(o)} = \underbrace{\text{KL}\bigl[ q(z|o) \mid p(z) \bigr]}_{\text{Complexity}} - \underbrace{\left(\mathbb{E}_{q(z|o)}\bigl[ p(o'|z) \bigr] \right)}_{\text{Accuracy}}
\end{equation}
It can be observed that minimizing FE is equivalent to maximizing ELBO. On the other hand, VAE can serve as a parametric implementation of predictive coding. The Weird World in Figure \ref{fig_pc_in_world_model} demonstrates the role of regularization loss or complexity in an exaggerated manner. When there exists a substantial discrepancy between the latent distribution mapped by the existing inference model (Encoder) and the distribution of newly observed data (the blue and red distribution curves in the figure), regularization loss or complexity can drive the model to converge rapidly. The prior \(p(z)\) is, in fact, empirical knowledge formed after a sufficient number of data samples have been observed, and it can be formally expressed as the marginal probability of the latent variable $z$.
\begin{equation}
    p(z) = \int p(o,z) do = \int p(o|z)p(z) do
\end{equation}
In the MWM, assuming that the observed objects are the actions of agents, and inferring their internal representations $z$ based on these actions—where $z$ may contain psychological variables such as \(\text{goal}\) and \(\text{belief}\)—the inference model can be expressed as:
\begin{equation}
    \text{Inference Model}: q_\phi(z|O_t^{ex})|p(O^{ex}_t|A_t^j)
\end{equation}
The above equation represents the inference of mental states based on the behaviors of agent $j$. Psychological states such as goal and belief are stored in $M_t$ and represented by $z$. By means of the generative model—referred to as Bayesian inverse planning in Bayesian ToM \cite{baker2017rational,baker2009action,shum2019theory}—the actions of agents can be inferred.
\begin{equation}
    \text{Generative Model}: p_\theta(\hat{A_t^{in}}|z)\label{eq_gen_model}
\end{equation}
The relationship between predictive coding and ToM lies in the fact that the former is a universal interpretive framework. Thus, ToM is a product of the PC mechanism that adopts Bayesian inverse planning as its computational model. In other words, predictive coding enables inference and generative representation of both physical and psychological phenomena within a unified model framework. Following the definition that a world model is an approximation of the environmental transition function \cite{richens2025general}, we can derive the transition function of the MWM using Equation \eqref{single_world_T} given the parametric model that infers actions from mental states as shown in Equation \eqref{eq_gen_model}. Therefore, the unity of the PWM and the MWM lies in the fact that both can be realized within the unified mathematical framework of predictive coding.

\subsection{The Necessity of Constructing a MWM}

\begin{figure}[!htb]
\centering
\includegraphics[width=1\textwidth]{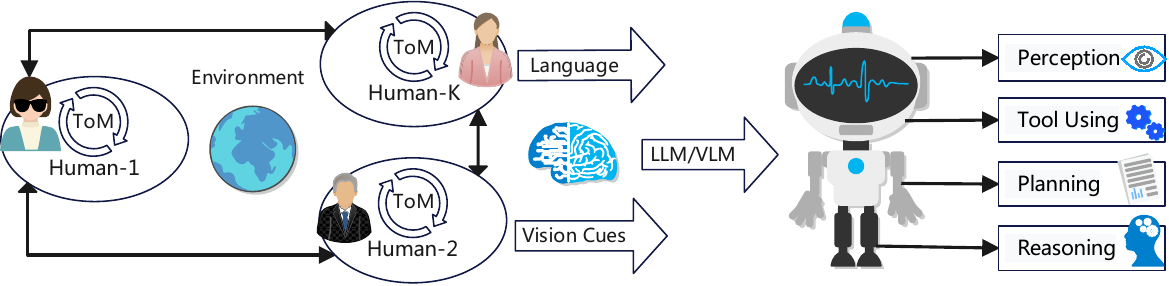}
\caption{Embodied AI Agent Framework based on LLM and VLM. Left: theory of mind is inherently performed in human society, the LLM and VLM are trained based on human language and vision cues from evironment. Right: the embodied AI agent is contructed based on LLM/VLM, which forms the main capability of reasoning and planning. }\label{fig_llm2agent}
\end{figure}

In Section \ref{section_physical_vs_mental}, we analyzed the differences between the PWM and the MWM. We concluded that the PWM only models the impact of physical actions on the evolution of the physical world, but fails to capture the process through which agents’ mental states influence environmental changes. From this analysis, we can initially establish the necessity of constructing the MWM. In this subsection, we further emphasize this necessity from both cognitive and computational modeling perspectives.
The theoretical prototype of the MWM can be traced back to explorations of human thinking mechanisms in the field of cognitive science. Developmental psychology research \cite{wellman1990child} has shown that humans’ understanding of the mind undergoes a gradual process, evolving from perceptual-goal psychology in infancy to false belief understanding in adulthood. In early infancy, humans can only predict behaviors based on observable physical events and direct goals (e.g., grasping an object in front of them) \cite{gergely1995taking}. Children aged 4-5 years old, however, begin to demonstrate the ability to understand others’ false beliefs through the Sally-Anne task \cite{baron1985does}. This landmark development lays the foundation for the formation of Theory of Mind (ToM) \cite{premack1978does}.
This theory reveals that humans possess the cognitive ability to construct "models of others’ mental states" and can distinguish between objective facts and subjective cognition. This capacity serves as an important theoretical basis for subsequent research on the MWM in the field of artificial intelligence.

In 2018, Rabinowitz et al. proposed the Machine Theory of Mind \cite{rabinowitz2018machine}. Humans do not rely on an understanding of the underlying neural mechanisms to comprehend other agents. Instead, they perform social reasoning through high-level mental models. Inspired by this, the authors suggested that machine learning systems could be trained to construct analogous models.
As illustrated in Figure \ref{fig_llm2agent}, the current mainstream architectures for embodied or virtual agents mostly adopt Large Language Models (LLMs) \cite{chang2024survey} and Vision-Language Models (VLMs) \cite{zhang2024vision} as their core "cognitive hubs". These foundation models serve as the core carriers of agents’ planning and reasoning capabilities. They can also expand their capacity boundaries through external tool invocation. Taniguchi et al. further pointed out in their research on population predictive coding \cite{taniguchi2024generative} that LLMs can be regarded as a concrete implementation of world models. The core rationale lies in the fact that human language is not merely a communication protocol. It is rather an externalized representation of a collective world model, which emerges from the decentralized, interactive meaning-making processes of socially embodied agents.
However, extensive benchmark test results show that sole reliance on LLMs tends to lead to shallow statistical pattern matching, making it difficult to achieve reliable recursive reasoning. Even state-of-the-art models such as GPT-4 \cite{openai2023gpt} and Qwen3 \cite{yang2025qwen3} still exhibit a significant performance gap compared with human-level performance on Theory of Mind-related benchmarks including BigToM \cite{gandhi2023understanding} and MuMA-ToM \cite{shi2025muma}.
Taniguchi et al. further divided world models into two categories \cite{taniguchi2024generative}. The first is the "internal model": a subjective model that an agent learns through its own sensorimotor interactions with the environment, which is used to predict future states and plan behaviors. The second is the "external model": an objective, structured knowledge representation of the world, entities, and their relationships, which is not tied to the direct experience of a single agent. These two categories correspond precisely to the dichotomy between Mental World Models (MWMs) and Physical World Models (PWMs). In addition, scholars such as Ding et al. also proposed that world models can be classified based on the type of knowledge they learn \cite{ding2025understanding}.
The aforementioned classification frameworks, combined with the empirical limitations of LLMs, jointly indicate that it is necessary to conduct specialized research and implementation on Mental World Models as an independent branch of world models.


\section{Element Representation of MWM}

\begin{table}[!htb]
    \centering
    \setlength{\belowcaptionskip}{0.2cm}
    \caption{Typical schools for representing mental world elements from the perspectives of psychology (strong representation) and computational neuroscience \cite{trappenberg2009fundamentals} (weak representation).}
    \scalebox{0.77}{
    \begin{tabular}{p{2.5cm}p{5cm}p{3cm}p{4cm}p{5.5cm}}
    \toprule
    \textbf{School Classification} & \textbf{Theoretical Perspective} & \textbf{Typical Representational Elements} & \textbf{Application Scenarios} & \textbf{Limitations}\\
 \midrule
 Folk Psychology \cite{dennett1989intentional} & Macroscopic approximate description: Assumes the mind consists of external-directed logical propositions & belief, desire, intention & BDI model agent design, motivation analysis & No "belief" entity in neuroscience; unable to explain irrational and unconscious behaviors. \\
 Structuralism \cite{freedheim2012handbook} & Attempts to decompose subjective experience into indivisible atoms via introspection & Sensations, Images, Affections & Basic psychophysics research, user interaction experience design & Unreliable introspection; "atomic" experiences vary across individuals and are difficult to standardize and verify. \\
 Evolutionary Psychology \cite{confer2010evolutionary} & The mind’s core consists of adaptive modules shaped by natural selection to solve ancient survival problems & Evolutionary adaptive modules, motivations, domain mechanisms & Criminal psychology, social psychology & Difficult to empirically validate the causal link between ancient environments and modern minds; weak explanatory power for individual differences. \\
 Psychoanalysis \cite{de2023ego} & The mind is a closed energy system, emphasizing repression, catharsis, and energy conservation & Id, Ego, Superego, Libido & Psychological counseling, advertising psychology & Difficult to experimentally verify the specific operation of the "unconscious" (i.e., unfalsifiable). \\
 Dimensional Emotion Theory \cite{russell1980circumplex} & Mental states are points in a continuous coordinate system rather than discrete switches & Valence, Arousal & Public opinion analysis, stress monitoring via wearable devices & Unable to distinguish complex emotions that are close in coordinates but distinct in nature (e.g., anger and fear). \\
 MidruleCognitive Architectures \cite{anderson1996act} & The mind is an information processing system; core elements are memory-stored data and CPU-executed rules & Knowledge Structures, Production Rules, Goal Stack, Working Memory & Human-computer interaction, cognitive modeling of complex tasks (e.g., driving) & Symbol Grounding Problem; difficulty handling ambiguity and creativity. \\
 Connectionism \cite{rumelhart1986parallel,rumelhart1986general} & Decentralized with no single "concept"; meaning resides in distributed network patterns & Weights, Activation Vectors & Deep learning (LLMs), pattern recognition & Poor interpretability; difficult to trace how a specific decision emerges from weights. \\
 Embodied Cognition \cite{barsalou1999perceptual,lakoff2024metaphors} & The mind is not confined to the brain but arises from body-environment interactions; representation is bodily simulation & Sensorimotor Schemas & Virtual reality interaction design, rehabilitation training & Difficult to explain the formation of disembodied abstract concepts such as "mathematics" and "justice". \\
 Predictive Coding / Free Energy Principle \cite{friston2010free,millidge2021predictive} & The brain is a prediction machine; core elements are not inputs but discrepancies between predictions and reality & Prediction Error, Priors & Computational psychiatry, active inference AI & Overly grand; explains everything but lacks specific predictive details. \\

\bottomrule
    \end{tabular}}

    \label{table_mental_world_element}
\end{table}

\begin{table}[!htb]
    \centering
    \setlength{\belowcaptionskip}{0.2cm}
    \caption{From the Perspective of Mental Reasoning Tasks: Basic Elements/Dimensions for Inferring Mental World States.}
    \scalebox{0.8}{
    \begin{tabular}{p{3cm}p{10cm}p{5.5cm}}
    \toprule
   \textbf{Taxonomy}&	\textbf{Representation Way} &  \textbf{Typical Method/Benchmark} \\
   \midrule
   Symbolic Belief	&Takes discrete symbolic texts and structured propositions as the core representational carriers of mental states, without architectural organization or probabilistic modeling.	&BigToM \cite{gandhi2023understanding}, ExploreToM \cite{sclar2024explore}, ToMLoc \cite{chen2025through}, MAgIC \cite{xu2024magic}, COKE \cite{wu2024coke}
\\
Probabilistic Belief	&Represents mental states in the form of probability distributions; the core is "neuro-guided online probabilistic assistance", which is completely different from the representational logic of symbolic/BDI architectures.	&NOPA \cite{puig2023nopa}
 \\
 Distributed Activation Vector& 	Enables the model to learn distributed activation vectors of "agent preferences (e.g., preferring item A over B)" and "goal intentions" in hidden layers by observing agents' "action sequences" (e.g., object retrieval in grid worlds).	& ToMNet \cite{rabinowitz2018machine}, LLM-Belief \cite{zhulanguage}, VToM \cite{chen2025through}
  \\
  BDI Architecture&	Organizes mental states around the BDI triad, with elements directly serving multi-agent collaborative decision-making.	&SoMi-ToM \cite{fan2025somi}, CoToMA \cite{li2023theory}, EmbodiedAI \cite{fung2025embodied}, ToM-Agent \cite{yang2025large}
   \\

\bottomrule
    \end{tabular}}

    \label{table_mental_world_element_datasets}
\end{table}

In physical world models, objective entities are typically characterized by observable physical properties, such as spatial position, geometry and materials at the macroscopic level, as well as mass, friction, stiffness and microscopic material structures at the more microscopic level. These properties jointly constitute the state space, which, combined with observation and transition dynamics, enables reasoning and decision-making under uncertainty (e.g., modeling paradigms represented by POMDP \cite{smallwood1973optimal,curtis2025llm}).
However, mental states or properties are not independent entities, but are represented by a set of abstract constructs in cognition and psychology. For instance, the Folk Psychology school \cite{georgeff1991modeling} holds that the basic elements of mental states are directed propositional attitudes such as Beliefs and Desires. Meanwhile, Psychoanalysis \cite{de2023ego} defines classic basic psychological elements including the Id, Ego, Superego, unconscious impulses and defense mechanisms. This paper does not propose a unique representation method for the MWM. Instead, it lists the dimensions of mental capabilities evaluated by relevant schools of psychology and computational neuroscience \cite{trappenberg2009fundamentals} (Table \ref{table_mental_world_element}) as well as related evaluation benchmarks (Table \ref{table_mental_world_element_datasets}). The purpose is to illustrate that in different application scenarios, corresponding mental states may be defined to characterize the mental world according to diverse application requirements and modeling concepts.

As shown in Table \ref{table_mental_world_element}, the representation of mental world elements constitutes a core topic at the intersection of psychology and computational cognitive neuroscience. Due to discrepancies in the priority of defining mental elements and the dependency characteristics of representational carriers, research paradigms in this field have evolved into two distinct categories: the strong representational paradigm in psychology, which centers on the explicit definition of mental units, and the weak representational paradigm in computational cognitive neuroscience, which prioritizes carriers and processes. These two paradigms jointly form the theoretical foundation for current interdisciplinary mental research, including the construction of artificial mental models.

The core logic of the strong representational paradigm lies in decomposing the mind into explicitly describable basic units, and achieving an approximate characterization of the mental world by defining the attributes and relationships of these units:
\begin{itemize}
\item Structuralism \cite{freedheim2012handbook} attempts to disassemble subjective experience into its "atomic units" through introspection (with sensations and affections as the core representational elements), providing an initial framework for the structural analysis of the mind. However, due to the subjectivity of introspection and the non-standardizability of experiential units, its methodology has gradually been replaced by Folk Psychology \cite{dennett1989intentional}. Folk Psychology adopts "external logical propositions" as the approximate representation of the mind (with belief, desire, and intention as core elements), circumventing the limitations of introspection and directly supporting practices such as BDI agent design and human motivation analysis. Nevertheless, its presupposition of the mind’s "rational propositional nature" makes it difficult to accommodate the non-rational behaviors described in neuroscience that involve "non-belief entities", thus exposing the inherent framework flaws of the strong representational paradigm.
\item Evolutionary Psychology \cite{confer2010evolutionary} and Psychoanalysis \cite{de2023ego} shift the research focus from "what the mind is" to "why the mind exists". Evolutionary Psychology represents the mind as "adaptive modules" shaped by natural selection (with evolutionary adaptive modules and survival/reproductive motivations as core elements), offering functionalist explanations for criminal psychology and social behavior analysis. Yet the link between ancient environments and modern minds lacks empirical validation. Psychoanalysis \cite{de2023ego}, on the other hand, defines the mind as a closed "energy system" (with ego and libido as core elements), focusing on energy dynamics such as repression and catharsis, and serving applications like psychological counseling and advertising-based psychological intervention. However, the non-embodiment of the "unconscious mind" has exposed it to persistent criticism regarding its "inability to verify specific operational mechanisms".
\item Dimensional Emotion Theory \cite{russell1980circumplex} breaks through the constraints of discrete units, representing mental states as continuous coordinate points such as valence and arousal, which enables the quantitative characterization of emotional states. This approach is well-suited for scenarios including public opinion analysis and stress monitoring via wearable devices. Its limitation, however, lies in the inability to distinguish between complex emotions that are "close in coordinate values but fundamentally different in nature" (e.g., the overlap of "high arousal/negative valence" between anger and fear), thus revealing the inherent flaw of "dimensional oversimplification" in the strong representational paradigm.
\end{itemize}

The core logic of the weak representational paradigm lies in downplaying the explicit definition of mental units, and instead focusing on the carrier forms or interaction processes of the mind. It indirectly achieves the representation of the mental world by characterizing the features of these carriers and processes:
\begin{itemize}
\item Cognitive Architectures \cite{anderson1996act} analogize the mind to an "information processing system", with knowledge structures, declarative chunks, and production rules as the core representational elements (corresponding to "memory data" and "CPU rules"). This approach has successfully supported cognitive modeling for complex tasks such as human-computer interaction and driving. However, due to the "symbol grounding problem" (the inability to explain the correspondence between symbols and the real world), it struggles to handle ambiguous and creative mental activities, which has become a core bottleneck in symbolic cognitive research.
\item Connectionism \cite{rumelhart1986parallel,rumelhart1986general} abandons the presupposition of "monolithic concepts" and represents mental meaning as a "decentralized network pattern" (with weights and activation vectors as core elements). The logic of its distributed activation vectors has directly empowered the practice of deep learning (e.g., LLMs) and pattern recognition. Nevertheless, because its decision-making process relies on "subsymbolic network activation", it has long been plagued by the problem of "extremely poor interpretability", making it difficult to trace the formation path of a single decision.
\item Embodied cognition and the predictive coding/free energy framework have further expanded the boundaries of representational carriers. Among them, Embodied Cognition \cite{barsalou1999perceptual,lakoff2024metaphors} posits that the mind is not confined to the brain, but is "a product of the interaction between the body and the environment" (with sensorimotor schemas as the core element). It has provided support for scenarios such as virtual reality interaction design and rehabilitation training, yet it fails to explain the formation of disembodied cognition such as mathematics and abstract concepts. The Predictive Coding/Free Energy Principle \cite{friston2010free,millidge2021predictive} defines the mind as a "prediction machine", where the core representational elements are not "input information", but rather "prediction errors" and "priors". Although it offers a new perspective for computational psychiatry (e.g., the interpretation of prediction errors in schizophrenia) and active inference AI, its excessive theoretical generalization—being able to "explain everything"—results in a lack of specific details regarding prediction regulation.
\end{itemize}

Research on mental state representation not only involves theoretical divergences at the epistemological level, but also has formed practical paradigms centered on technical carriers and element organization logic from a methodological perspective. These paradigms directly correspond to the mainstream methods and evaluation benchmarks in the current field of mental reasoning (as shown in Table \ref{table_mental_world_element_datasets}), reflecting the specific pathways for the transformation of cognitive theories into technical implementations. Based on the discreteness of representational carriers, uncertainty quantification methods, and architectural organization logic, the methodological domain can be divided into four major paradigms: symbolic belief, probabilistic belief, distributed activation vector, and BDI architecture. Their core characteristics, typical applications, and academic values are as follows:
\begin{itemize}
\item Symbolic belief takes discrete symbolic texts and structured propositions as the core representational carriers of mental states. It downplays the architectural correlations or uncertainty quantification of elements, and achieves reasoning solely through the combination of symbolic units. This paradigm represents the direct technical implementation of the "strong representational schools" from an epistemological perspective (e.g., the propositional mind in Folk Psychology and the experience decomposition in Structuralism). The propositional descriptions of "belief and desire" in Folk Psychology are transformed into textual symbols such as "Agent A believes X" in benchmarks including BigToM \cite{gandhi2023understanding} and ExploreToM \cite{sclar2024explore}. The "experience atoms" in Structuralism, on the other hand, correspond to the structured knowledge graph symbols of "entity-attribute-state" in COKE \cite{wu2024coke}.
\item Probabilistic belief employs probability distributions as the carriers of mental states, realizing dynamic reasoning by quantifying uncertainty (e.g., "there is an 85\% probability that the object is on the table"). Its logic is completely heterogeneous from that of discrete symbols and architectural organization. This paradigm corresponds to the technical practice of the "Predictive Coding/Free Energy Principle" from an epistemological perspective. The "discrepancy between prediction and reality" in predictive coding is translated into "neuro-guided online probabilistic assistance" in NOPA \cite{puig2023nopa}, which updates the probability distribution of demand beliefs in real time based on user behavior.
\item Distributed activation vector uses distributed activation vectors in the hidden layers of neural networks as the carriers of mental states. It indirectly characterizes the mind through features such as vector clustering and similarity (e.g., activation clusters in the layers of LLMs correspond to the beliefs of different agents), with no explicit discrete symbols. For instance, Zhu et al. \cite{zhulanguage} manipulated the identified neural representations directly during the reasoning process to establish a causal relationship between distributed activation vectors and social reasoning. They guided activations along specific directions to test whether these internal representations would functionally affect the model’s ToM reasoning performance, a method referred to as linear probing. Through linear probing technology, decodable belief representations were identified in the attention head activations of models such as Mistral-7B-Instruct and DeepSeek-LLM-7B-Chat. In addition, the identified directions of belief representations across different social reasoning tasks exhibited high consistency, with the correlation coefficient of accuracy rates reaching 0.85–0.90. This indicates that these neural representations may possess cross-task generalization capabilities.
\item The BDI architecture organizes mental elements around the Belief-Desire-Intention (BDI) triad, and the correlations between elements directly serve the collaborative decision-making of multi-agent systems (e.g., deriving execution intentions from collaborative goals). The representation type encompassing belief, intention, goal, and emotion proposed by Fung et al. \cite{fung2025embodied} in 2025 does not fall entirely under the BDI architecture, but can be categorized as an "extended BDI architecture". Mozikov et al. \cite{mozikov2024eai} pointed out that existing evaluations of "safety and human alignment" for LLMs mostly rely on pure natural language benchmarks, which have significant limitations. Human decision-making is often driven by emotions (e.g., anger and joy can alter choices), yet previous studies have not systematically explored the impact of emotions on the decision-making logic and ethical tendencies of LLMs. Even aligned LLMs may exhibit irrational behaviors due to emotional biases (e.g., deception and a sharp drop in cooperation rates), which pose risks for scenarios requiring autonomous decision-making such as medical care and customer service.
\end{itemize}

A review of these four methodological paradigms reveals that they all revolve around a core trade-off between interpretability and flexibility. Symbolic belief and the BDI architecture prioritize explicit interpretability of mental representation at the expense of flexibility in scenario adaptation. The former anchors mental states via discrete texts or structured propositions, while the latter clarifies the correlation logic of collaborative elements through a triadic architecture. Both allow researchers to directly trace the symbolic pathways of reasoning. In contrast, distributed activation vectors and probabilistic belief place greater emphasis on flexibility in scenario adaptation, albeit at the cost of reduced transparency in the representation process. The former characterizes complex mental states by leveraging vector patterns in the hidden layers of neural networks, while the latter quantifies uncertainty in dynamic scenarios through probability distributions. However, the reasoning processes of both are difficult to decompose into intuitive symbolic units.
Notably, these methodological paradigms are not isolated technical approaches. In recent years, cutting-edge research has sought to move beyond this binary choice. For example, some studies have attempted to integrate symbolic belief with distributed activation vectors, extracting interpretable symbolic belief units from the hidden-layer activation vectors of large language models via linear decoding techniques. The essence of such integration lies in attempting to bridge the inherent contradiction between interpretability and flexibility at the methodological level, building on the epistemological theoretical foundation. This direction is gradually emerging as a core focus of exploration in the field of mental state representation.


\begin{table}[!htb]
    \centering
    \setlength{\belowcaptionskip}{0.2cm}
    \caption{Related methods for ToM reasoning
Among these methods, ToM Prompting is essentially a form of context engineering, while Model-based Inference essentially relies on human-guided reasoning chains. Lev. means ToM levels.}
    \scalebox{0.8}{
    \begin{tabular}{p{4em}p{2cm}p{1cm}p{1cm}p{3.5cm}p{3.5cm}p{5.5cm}}
    \toprule
   \textbf{Taxonomy}	&\textbf{Method}	&\textbf{Year}	& \textbf{Lev.} &\textbf{Base LLM} & \textbf{Reasoning Paradigm} &\textbf{Key Mechanism}	\\
   \midrule
\multirow{10}{4em}{ToM Prompting} 
& Generative Agent \cite{park2023generative}& 2023 & 1 & GPT-3.5-turbo&  Retrieval-Enhanced Neural Language Model Reasoning & Retrieval-Based Storage and Recursive Reflection Generation of Memory Streams   \\
& CoT-ToM \cite{moghaddam2023boosting} &2023& 1 & GPT-4/GPT-3.5&Language-Guided Stepwise Reasoning & CoT prompting, In-context learning  \\
&CoToMA \cite{li2023theory}	& 2023	&3 &GPT-4/GPT-3.5&Neural language-based reasoning&To structurally infer and understand others' perspectives and intentions	\\
&SymbolicToM \cite{sclar2023minding}	&2023	&2 & GPT-4/Llama-13B&Neuro-symbolic hybrid&Symbolic belief graph tracking, Witness-based knowledge propagation	\\

&SimToM \cite{wilf2024think}		&2024&2 &GPT-4/Llama2-13B&Simulation-based reasoning&
Perspective-taking filtering, Two-stage prompting	\\
& COKE \cite{wu2024coke} & 2024 & 1&  Llama-2-7B/Mistral-7B& Neural-Symbolic Fusion&Cognitive Knowledge Graph, Chained Cognitive Reasoning   \\

& MindForge \cite{licuamindforge} & 2025 & 2&GPT-4/Llama-3.1-8B& Neural-Symbolic Fusion + Casual Reasoning&Natural language inter-agent communication, ToM causal template  \\

&ToM-Agent \cite{yang2025large} & 2025 &2 & GPT-4/GPT-3.5 & Neuro-symbolic + Simulation-based reasoning& BDI tracking with confidence disentanglement, Counterfactual reflection\\
&XToM \cite{chan2025xtom} & 2025 & 2 &GPT-4o/DeepSeek R1&Neural language-based reasoning& Cross-Language Consistency Evaluation \\

&DEL-ToM \cite{wu2025tom} & 2025 & 4& GPT-4o/Llama3.1-8B &Neural-Symbolic Fusion& PBM-based trace verification, inference-time scaling \\
& VToM \cite{chen2025through} & 2025 & 1 & GPT-4o &Multimodal neural reasoning & Key frames retrieval\\
\midrule
\multirow{10}{4em}{Model-based Inference}
&ToMNet \cite{rabinowitz2018machine} & 2018 & 1& -& Implicit Reasoning via Neural Networks & character, mental state, and prediction networks  \\
&PGM-Aware Agent \cite{xu2024magic}		&2024	& 3 & GPT-o1/Llama-2/Claude-2& Neuro-Symbolic Fusion + Probabilistic Reasoning&PGM-LLM fusion, Two-hop understanding	
\\
&BIP-ALM \cite{jin2024mmtom}		&2024&1&GPT-4/Video-Llama 2	&Neural-Symbolic Fusion + BIP&Bayesian inverse planning, Language model-accelerated likelihood estimation\\
& BToM-EL \cite{ying2024grounding} & 2024 &1&unknown&Neural-Symbolic Fusion + BIP& Bayesian Inverse Planning，
Cognitive Logic Evaluation，
Natural Language-Logic Conversion  \\
&LIMP \cite{shi2025muma}	&2025&2&GPT-4o/Gemini 1.5 Pro&Neural-Symbolic Fusion + BIP&	Bayesian Probabilistic Inference	\\

& Thought-tracing \cite{kim2025hypothesis} & 2025 &2&CPT-4o, DeepSeek R1, Qwen2.5& Neural-Symbolic Fusion + Sequential Monte Carlo&Hypothesis generation and propagation, Action likelihood-based weighting	\\
& MetaMind \cite{zhang2025metamind} & 2025 &2&GPT-4, Claude-3.5, DeepSeek V3/R1 & Neural-Symbolic Fusion + BIP&Human-like Social Reasoning via a Three-stage Metacognitive Cycle  \\
&AutoToM \cite{zhang2025autotom}	&2025&	Any&GPT-4o, Llama 3.1 70B, Gemini 2.0& Neural-Symbolic Fusion + BIP&Automated Bayesian Inverse Planning and agent model discovery	\\
\bottomrule
    \end{tabular}}

    \label{table_ToM_models}
\end{table}

\section{Theory of Mind: from Static Representation to Dynamic Reasoning}
ToM serves as the operational mechanism or inference engine of the MWM. Its core function is to infer unobservable hidden mental states (e.g., the belief that "he thinks the cup is in the drawer" or the goal that "he wants to drink water") from observable behaviors (such as language, actions, and facial expressions) \cite{frith2005theory, wellman2018theory}.
From a fundamental philosophical perspective, ToM is a modern mathematical variant of Folk Psychology/Intentional Stance \cite{dennett1989intentional}. However, in terms of implementation methods, it often incorporates Connectionism \cite{rumelhart1986parallel, rumelhart1986general} (for perceptual processing) and Cognitive Architectures \cite{anderson1996act} (for reasoning).

\begin{figure}[!htb]
\centering
\includegraphics[width=1.0\textwidth]{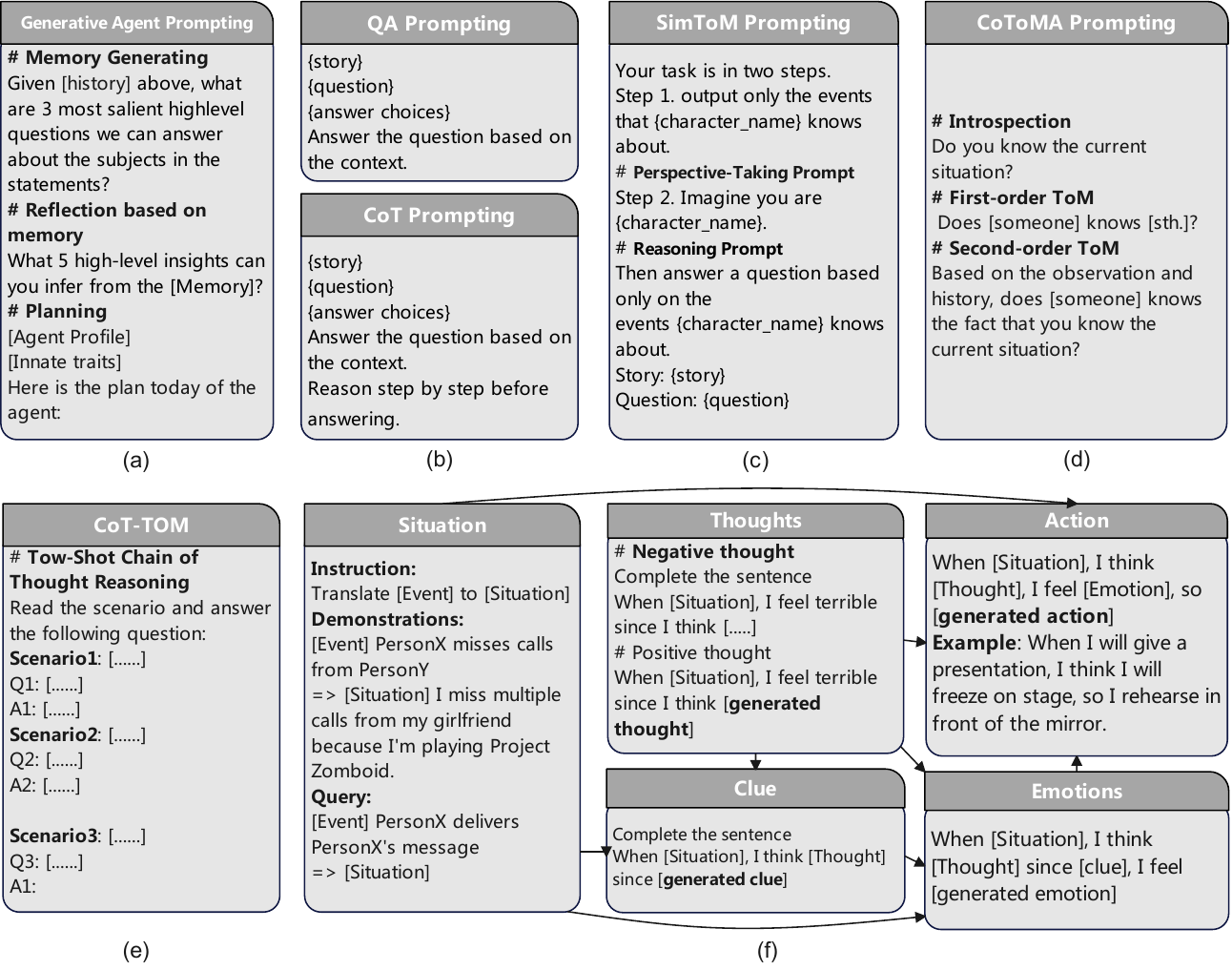}
\caption{Typical Prompts of ToM Prompting Methods.
}\label{fig_prompting_examples}
\end{figure}


\subsection{Prompting Paradigm: Stimulating the Implicit ToM Capabilities of LLMs}

In recent years, large language models (LLMs) have demonstrated remarkable capabilities in natural language understanding and generation tasks. Researchers have found that through specific prompt design (i.e., "context engineering") \cite{sahoo2024systematic}, the implicit Theory of Mind (ToM) reasoning abilities of LLMs can be elicited, thus forming the "prompting paradigm" for ToM modeling \cite{bubeck2023sparks}, as illustrated in Table \ref{table_ToM_models}.
The core idea of this paradigm is as follows: without the need for additional parameter fine-tuning of LLMs, the model can be guided to perform reasoning in accordance with ToM logic solely by incorporating contextual information such as "perspective guidance" and "reasoning step prompts" into the input \cite{zhu2024language}. Its theoretical foundation stems from the Simulation Theory in cognitive science—namely, understanding others’ mental states by "simulating their cognitive processes" \cite{goldman1989interpretation,goldman2006simulating}.

Under the Prompting paradigm, the pioneering representative work that first achieved first-order ToM capabilities is Generative Agent \cite{park2023generative}. By constructing a persistent memory management and retrieval mechanism, it enables agents to track and reason about others’ knowledge states, beliefs, and intentions. As illustrated in Figure \ref{fig_prompting_examples}(a), its core lies in three key steps:
\begin{itemize}
    \item Utilizing memory streams to store all perceptual experiences, and dynamically extracting behavioral records of others relevant to the current context through a three-dimensional weighted retrieval mechanism based on recency, importance, and relevance.
    \item Recursively synthesizing low-level observations into high-level reflections (e.g., "Sam may not know about the party"), thereby forming explicit inferences about others’ mental states.
    \item Generating socially coordinated behaviors (e.g., actively spreading information, extending invitations) based on these inferences.
\end{itemize}
This architecture is implemented on GPT-3.5-turbo, integrating neural language models with a retrieval-augmented reasoning paradigm. In the Smallville sandbox environment, it demonstrated typical first-order ToM behavioral manifestations such as information dissemination, relationship maintenance, and collective activity coordination. However, the ToM capabilities of this method are limited to first-order reasoning, failing to exhibit second-order or higher-order nested mental state inferences (e.g., "A thinks B believes that C does not know X"). Its interpretability presents partially transparent characteristics: the reflection module labels the sources of evidence (e.g., "Based on memories 1, 5, 3") and the memory retrieval process is traceable, but the final behavioral decision-making relies on the black-box generation of large language models, lacking an explicit symbolic reasoning chain.

To address the limitations of Generative Agent in explicating the reasoning chain, CoT-ToM \cite{moghaddam2023boosting} further optimizes the prompting strategy and proposes a prompt engineering method based on language-guided stepwise reasoning, aiming to improve the performance of large language models (LLMs) in first-order Theory of Mind (ToM) reasoning. Without the need for additional training or fine-tuning, this method significantly enhances the performance of GPT-4 and GPT-3.5 series models on false-belief tasks through carefully designed prompting strategies. Its core mechanism consists of three collaborative components (Figure \ref{fig_prompting_examples}(b) illustrates the differences between direct question answering and Chain-of-Thought (CoT) reasoning):
\begin{itemize}
    \item Two-shot CoT prompting (as shown in Figure \ref{fig_prompting_examples}(e)), which guides the model to mimic the reasoning pattern by presenting examples that incorporate intermediate reasoning steps.
    \item Step-by-step instructions, which explicitly require the model to decompose the reasoning process.
    \item In-context learning, which integrates the above two components to maximize reasoning quality.
\end{itemize}
This method adopts a natural language reasoning paradigm, where the model outputs a complete reasoning chain structured as "fact identification $\rightarrow$belief state tracking$\rightarrow$logical inference → conclusion", demonstrating high interpretability. Specifically, each reasoning step is traceable and verifiable, with explicit justifications provided especially for critical knowledge asymmetry judgments (e.g., "Anne does not know because she was not present at the time"). Experimental results show that prompt optimization alone improves the accuracy on first-order ToM tasks from 40–80\% in the zero-shot setting to 80–100\% (with GPT-4 achieving a perfect score), surpassing the 87\% human baseline. This study reveals that the ToM capabilities of LLMs are context-dependent—the models inherently possess potential reasoning abilities but require appropriate guidance to express them. It also validates the effectiveness of interpretable reasoning chains for complex cognitive tasks, laying a solid foundation for advancing the application of LLMs in social understanding and commonsense reasoning.

As a typical practice of the Prompting paradigm, SimToM \cite{wilf2024think} has proposed a design that is more aligned with human cognition in terms of perspective simulation. It constructs a two-stage prompting framework to mimic the human ToM reasoning process. As illustrated in Figure \ref{fig_prompting_examples}(c):
\begin{itemize}
    \item The first stage is the Perspective-Taking Prompt, which guides the model to switch to the perspective of the target agent (e.g., "Now please act as Anna, recall where you placed the apple earlier, and note that you are unaware that John moved the apple").
    \item The second stage is the Reasoning Prompt, which requires the model to answer ToM questions based on this perspective (e.g., "As Anna, where will you look for the apple first?").
\end{itemize}
The innovation of SimToM \cite{wilf2024think} lies in the following: it simulates the role-taking process in human ToM reasoning via "perspective impersonation", rather than directly requiring the model to reason from a third-party perspective, which is more consistent with human cognitive mechanisms. Experimental results show that on the BigToM \cite{gandhi2023understanding} benchmark, SimToM \cite{wilf2024think} achieves a 22.9\% accuracy improvement over direct QA (e.g., "Where will Anna look for the apple first?") and a 7.5\% improvement over chain-of-thought reasoning \cite{wei2022chain} through its two-stage prompting strategy.
However, the limitations of SimToM \cite{wilf2024think} are also quite pronounced. First, it is heavily reliant on the implicit knowledge of LLMs—if LLMs have not acquired ToM concepts such as "false belief" through pre-training, performance improvement will be difficult to achieve even with prompting. Second, its high-order reasoning capability is fragile—in third-order and higher-order ToM tasks, the effectiveness of the two-stage prompting drops significantly, and the model is prone to perspective confusion (e.g., misinterpreting "Anna thinks Bob believes" as "Bob believes"). Finally, it lacks the ability to process multimodal information—SimToM \cite{wilf2024think} is only applicable to textual scenarios, unable to integrate multimodal cues such as visual and action information, making it difficult to support multimodal interaction for embodied agents.

When the Prompting paradigm is extended from single-agent scenarios to multi-agent collaboration scenarios, CoToMA \cite{li2023theory} emerges as a pivotal exploration. It elicits the joint reasoning capability of large language models (LLMs) regarding the mental states of multiple agents by designing multi-agent dialogue prompts. For instance, in a "bomb disposal task" scenario—where multiple agents must collaborate to defuse a bomb, with each agent possessing partial information—the prompt of CoToMA \cite{li2023theory} would include the following: "Agent A knows that the bomb password is 123, but does not know whether Agent B is aware of it. Agent B knows that the bomb has a 5-minute countdown, but does not know whether Agent A knows the password. Now, please act as a coordinator to infer the respective beliefs of Agent A and Agent B and design a collaboration strategy." By simulating information asymmetry and collaboration requirements among multiple agents, CoToMA \cite{li2023theory} tests the model’s ability to understand collective mental states.
In terms of mental reasoning evaluation, this dataset designs three levels of reasoning tasks: introspection, first-order Theory of Mind (reasoning about others’ belief states), and second-order Theory of Mind (reasoning about others’ perceptions of one’s own beliefs), as illustrated in Figure \ref{fig_prompting_examples}(d). However, such methods still fail to break through the inherent limitations of the Prompting paradigm. First, symbolic representation requires manual design—different scenarios necessitate the definition of distinct symbolic systems (e.g., belief symbols, goal symbols), resulting in poor generalizability. Second, multi-agent collaborative reasoning still relies on textual dialogue, making it unable to handle physical interactions among embodied agents (e.g., movement coordination between agents). Finally, the reasoning capability of the model is still constrained by the pre-training data of LLMs; if the scenario falls outside the scope of pre-training coverage (e.g., emerging social dynamic scenarios), performance will decline significantly.

Beyond prompting methods that rely on the implicit knowledge of large language models (LLMs), neuro-symbolic fusion also provides a structured pathway for ToM reasoning, with COKE \cite{wu2024coke} serving as a representative implementation of this approach. It proposes a first-order Theory of Mind reasoning framework based on cognitive knowledge graphs, formalizing human social cognitive processes into learnable structured knowledge via the neuro-symbolic fusion paradigm. As illustrated in Figure \ref{fig_prompting_examples}(f), the core innovation of this method lies in instantiating ToM capabilities into 45,369 manually validated cognitive chains. Each chain explicitly models human mental activities and their behavioral/emotional responses in specific social scenarios through a five-node structure: "Situation $\rightarrow$ Cue $\rightarrow$ Thought $\rightarrow$ Action + Emotion". This chain-based representation not only preserves the causal logic of ToM reasoning but also characterizes the diversity of cognitive processes through polarity labeling (positive/negative).
In terms of implementation, COKE adopts a two-stage strategy. First, it leverages GPT-3.5 to generate candidate data, which is then filtered and revised by psychology professionals to construct a high-quality knowledge graph. Second, it trains the Cognitive Language Model (COLM) on LLaMA-2 via multi-task learning, enabling it to generate complete cognitive chains for unseen scenarios. This method significantly outperforms strong baselines such as GPT-4 across four cognitive generation tasks (cue, thought, action, emotion), while also demonstrating high interpretability—each reasoning path can be traced back to specific nodes, providing a reliable ToM capability injection mechanism for downstream social applications (e.g., emotional support dialogue). Its limitations include a restriction to first-order ToM reasoning and a narrow thematic scope (covering only five categories of social topics), making it unable to handle high-order nested reasoning scenarios such as "A thinks B believes C".

To address the core pain points of reasoning opaqueness and insufficient robustness in the Prompting paradigm, SymbolicToM \cite{sclar2023minding} introduces symbolic representation and develops a symbol-enhanced Prompting method. Its core design lies in embedding a symbolic belief tracker into prompts—by defining structured symbols (e.g., "Belief (Agent, Object, Location, Time)"), it transforms ambiguous natural language reasoning into explicit symbolic logic reasoning. For instance, in the false-belief scenario of "Alice and Bob", the prompt of SymbolicToM \cite{sclar2023minding} would include: "Belief (Alice, Book, Shelf A, \(t_1\)) → Event: Bob moves the book to Shelf B at \(t_2\) (unobserved by Alice)$\rightarrow$ Belief (Alice, Book,?, \(t_3\))", guiding the model to infer via symbolic logic that "Alice’s belief at \(t_3\) is Shelf A".The advantages of such symbolic enhancement are twofold:Improved reasoning interpretability: Every step of symbolic logic reasoning is traceable, overcoming the opaqueness issue of text-only prompts.Enhanced robustness: The model is less susceptible to variations in textual expressions (e.g., replacing "move to" with "place on"), since symbolic representations are decoupled from specific wordings.Mental reasoning requires not only modeling beliefs but also integrating emotions and social norms to handle complex interactions. MAgIC \cite{xu2024magic} and Hi-ToM \cite{wu2023hi} explore this direction from the perspectives of emotion-driven inference and social norm comprehension, respectively. Specifically, MAgIC \cite{xu2024magic} verifies agents’ ability to infer emotion-driven strategies through multi-agent game scenarios—for example, identifying an opponent’s tendency to "betray" triggered by the emotion of "anger", which demonstrates the significance of emotional modeling for interactive strategy selection. Hi-ToM \cite{wu2023hi} focuses on social norm understanding and deceptive intention recognition, designing an experiment of "separation of public and private communication" to test agents’ capability to detect deception. In collaborative tasks, robots must judge the nature of misleading information from partners and choose between "correcting" or "guarding against" strategies, which illustrates that understanding social roles, rules, and interpersonal relationships is an essential prerequisite for parsing complex interactions.

Most of the aforementioned methods focus on first-order ToM or disembodied scenarios, while MindForge \cite{licuamindforge} breaks through this limitation by proposing an embodied agent collaborative learning framework centered on second-order ToM. This approach transcends the constraints of traditional isolated single-agent learning, integrating cultural learning theory with explicit ToM modeling to enable open-source LLM-driven agents to achieve performance comparable to that of GPT-4.
In terms of ToM reasoning mechanisms, MindForge adopts a neuro-symbolic fusion paradigm and constructs a structured BDI representation system based on the causal templates of BigToM \cite{gandhi2023understanding}. Each agent not only maintains its own perceptual beliefs, task beliefs, and interaction beliefs but, more crucially, explicitly models the complete mental states of collaborative partners (partner beliefs), thereby realizing second-order mental reasoning. This hierarchical belief architecture connects perception and action via causal chains (percepts$\rightarrow$beliefs$\rightarrow$desires$\rightarrow$actions) and dynamically updates partner models through multi-round natural language dialogues, allowing agents to perform perspective-taking—expert agents can diagnose the false beliefs of novices and provide targeted guidance, while novices can ask precise questions by simulating the mental states of experts.
The high interpretability of this method is reflected in three aspects: all belief states are explicitly represented in symbolic form. The reasoning process unfolds along traceable causal graphs. Dialogue records fully preserve interaction traces. Experimental results show that in the Minecraft environment, after only 7 rounds of collaborative dialogue, MindForge agents powered by Mixtral-8x7B increased their success rate on basic tasks from 29\% to 79\%, and the number of tech tree milestones reached 3 times that of the Voyager baseline. These findings verify the core value of second-order ToM reasoning in embodied collaborative learning—achieving the social transmission and accumulation of knowledge by explicitly modeling "how others think".

In the dynamic scenario of open-domain dialogue, ToM-Agent \cite{yang2025large} optimizes the mechanism for tracking the interlocutor’s mental states by decoupling beliefs and confidence levels. Its core lies in constructing the ToM-agent paradigm, which enables generative agents to simultaneously maintain their own BDI (Belief, Desire, Intention) representations and top-k inferences of the interlocutor’s BDI during open-domain dialogue, while updating confidence levels via a foresight-reflection cycle: the agent first predicts the interlocutor’s next-turn response based on the currently inferred BDI, then compares the predicted response with the actually observed response, and indirectly evaluates the accuracy of BDI inference through counterfactual reasoning, thereby optimizing the confidence update strategy.
This method primarily addresses the problem of tracking unobservable mental states in open-domain dialogue, with a particular focus on resolving conflicts between personalized beliefs and commonsense knowledge. It has been validated on two benchmark dialogue datasets, EmpatheticDialogues and PersuasionForGood, and demonstrates outstanding performance on both first-order ToM tasks (inferring the interlocutor’s BDI) and second-order ToM tasks (inferring whether the interlocutor understands one’s own BDI). However, its performance is subject to three key limitations:
\begin{itemize}
    \item It may be constrained by the inherent commonsense biases of large language models, which hinder the modeling of personalized beliefs that contradict commonsense knowledge.
    \item It requires manual annotation to evaluate the quality of BDI inference.
    \item It faces decision optimization issues where dialogues may terminate prematurely when confidence levels fail to reach a threshold value.
\end{itemize}

Small language models often exhibit poor performance in high-order ToM reasoning due to parameter scale constraints, while DEL-ToM \cite{wu2025tom} addresses this issue via inference-time expansion rather than architectural modifications. The core innovation of this method lies in combining neural reasoning with symbolic verification, which proceeds in two key steps:
\begin{itemize}
    \item First, it formalizes ToM tasks as a verifiable sequence update process of belief states using Dynamic Epistemic Logic (DEL). Through the product update mechanism of DEL, it accurately models the belief evolution of multi-agent systems.
    \item Second, it trains a Process Belief Model (PBM) as a verifier, which is supervised and trained using process-level labels automatically generated by a DEL simulator, thereby acquiring the capability to evaluate the reliability of intermediate reasoning steps.
\end{itemize}
In the inference phase, the model generates multiple candidate belief trajectories, and the PBM scores each step of every trajectory. The optimal path is then selected via Best-of-N or Beam Search strategies. Experimental results show that this method significantly improves the performance of small models on 0–4th order ToM tasks (e.g., the average accuracy of Llama3.2-3B is increased by 33.6\%). Moreover, the PBM can be transferred across models without retraining. Compared with the GRPO fine-tuning method that requires substantial computational resources, DEL-ToM only needs 3 hours to complete PBM training on a single GPU, without modifying the parameters of the base model, thus avoiding the risk of performance degradation on other tasks. Its high interpretability stems from the formal representation provided by DEL, which endows each belief update step with symbolic semantics and verifiability, laying a foundation for trustworthy socially reasoning AI.

Beyond monolingual scenarios, cross-lingual consistency is also a key metric for evaluating whether ToM capabilities constitute genuine cognitive reasoning, and XToM \cite{chan2025xtom} conducts an in-depth analysis of this aspect. It adopts a neural reasoning paradigm and tests two prompting strategies: zero-shot and CoT. Specifically, CoT guides the model to generate reasoning steps via the prompt "let's think step by step", which provides limited interpretability but does not qualify as explicit symbolic reasoning. Further cross-lingual consistency analysis shows that models yield highly consistent cross-lingual outputs on factual questions, yet exhibit language dependence in belief reasoning. This finding reveals that the ToM capabilities of models may essentially rely on statistical pattern matching rather than language-independent cognitive reasoning. In addition, transfer learning experiments confirm that multi-task joint fine-tuning can enhance the cross-lingual generalization of ToM capabilities. While monolingual fine-tuning also produces cross-lingual transfer effects, multi-lingual joint fine-tuning delivers more robust performance.
Text-only ToM reasoning struggles to capture visual cues in dynamic social interactions, whereas VToM \cite{chen2025through} extends ToM reasoning to the video multimodal domain, enabling mental state inference in dynamic scenarios. Targeting first-order ToM tasks, this method is built on the Video-ChatGPT architecture and designs a frame relevance scoring mechanism to locate video clips that carry key mental cues. It achieves cross-modal reasoning through joint alignment of visual encoders and text embeddings. Unlike traditional text-based ToM research that relies on static linguistic cues, VToM leverages temporal visual signals such as facial expressions and body language to capture the dynamic evolution of beliefs and intentions. On the Social-IQ 2.0 dataset, it raises the question-answering accuracy to 47.15\% and the frame localization accuracy to 77.97\%. Its reasoning paradigm falls into the category of end-to-end multimodal neural reasoning, providing a moderate level of interpretability via retrieved key frames—though it does not output explicit reasoning chains, the localization results reveal the spatiotemporal evidence focused on by the model, thus offering visual traceability for understanding how AI "reads minds". This work highlights the necessity of the video modality for complex ToM reasoning and provides a new paradigm for building AI systems with social intelligence.

The strengths of ToM Prompting-related methods lie in the progressive refinement of the paradigm. Starting from text-only prompting, these methods integrate neuro-symbolic techniques and symbol enhancement mechanisms, achieving capability expansion from first-order to second-order ToM, from single-agent to multi-agent scenarios, and from text to multimodal settings. Their applicable scenarios have also extended from static reasoning to dynamic social interaction and embodied interaction.
Meanwhile, their interpretability has been continuously improved, evolving from partially transparent reflection to explicit reasoning chains and symbolic logic verification. Certain methods can trace the reasoning process via knowledge graphs or causal graphs. Moreover, most of them do not require large-scale fine-tuning; instead, they balance performance and resource efficiency through prompt design or lightweight training, with some components even featuring cross-model transferability.
However, their limitations are equally prominent:
\begin{itemize}
    \item Weak high-order reasoning capability: Most methods perform poorly in high-order reasoning tasks, with only a few covering second-order and above reasoning. Perspective confusion is prone to occur in high-order tasks.
    \item Insufficient depth of multimodal fusion: Only a limited number of approaches have attempted video modality integration, failing to achieve deep multimodal fusion, which makes it difficult to support embodied physical interactions.
    \item Poor generalizability: Performance relies heavily on the pre-trained knowledge of LLMs and manual symbol design, resulting in weak adaptability to unseen scenarios or cross-scenario applications.
    \item Narrow scenario and topic coverage: Most methods focus on specific interactions or a small number of social topics, making them unable to address the complex and ever-changing ToM demands of the real world.
\end{itemize}

\subsection{Model-Based Inference Paradigm: Constructing Interpretable Mental Models}

Unlike the Prompting paradigm, which relies on the implicit capabilities of large language models (LLMs), the core idea of the Model-Based Inference paradigm is to "explicitly construct mental models"—that is, to simulate the cognitive mechanisms of human Theory of Mind (ToM) through well-defined mathematical frameworks (e.g., Bayesian inverse planning, neuro-symbolic fusion frameworks), thereby achieving interpretable and generalizable reasoning about mental states.The advantages of this paradigm are as follows: the reasoning process is transparent and controllable, and it can adapt to different scenarios through model adjustments, making it more suitable for the long-term cognitive needs of embodied agents. Moreover, when LLMs are incorporated to assist in complex decision-making, it can evaluate the probability of action \(a^t\) based on structured belief assumptions (including environmental state \(s^t\), goal \(g^t\), and historical belief \(b^{t-1}\)) \cite{huang2022language,li2022pre}. Table \ref{table_ToM_models} lists the relevant research works of this category in recent years, and Table \ref{table_diagram_compare} presents a comparison between the Model-Based Inference and ToM Prompting paradigms.

\begin{table}[!htb]
    \centering
\caption{Qualitative Comparison of the Prompting and Model-Based Inference Paradigms}
\scalebox{0.9}{
\begin{tabular}{p{4cm}p{7cm}p{7cm}}
\toprule
Evaluation Dimension & Prompting Paradigm & Model-Based Inference Paradigm \\
\midrule
Implementation Complexity & No fine-tuning required, only prompt design needed & Requires construction of mathematical models and reasoning frameworks \\
Interpretability & Black-box reasoning relying on the implicit capabilities of LLMs & Explicit models with traceable reasoning processes \\
Robustness of High-Order Reasoning & Accuracy drops significantly in third-order and higher-order reasoning tasks & Can handle high-order scenarios via recursive reasoning frameworks \\
Generalizability & Relies on pre-trained data with poor performance on novel scenarios & Can adapt to new scenarios through model adjustments \\
Real-Time Performance & Fast inference speed of LLMs & High computational complexity (e.g., Bayesian reasoning) \\

\bottomrule
    \end{tabular}}

    \label{table_diagram_compare}
\end{table}

\begin{figure}[!htb]
\centering
\includegraphics[width=1.0\textwidth]{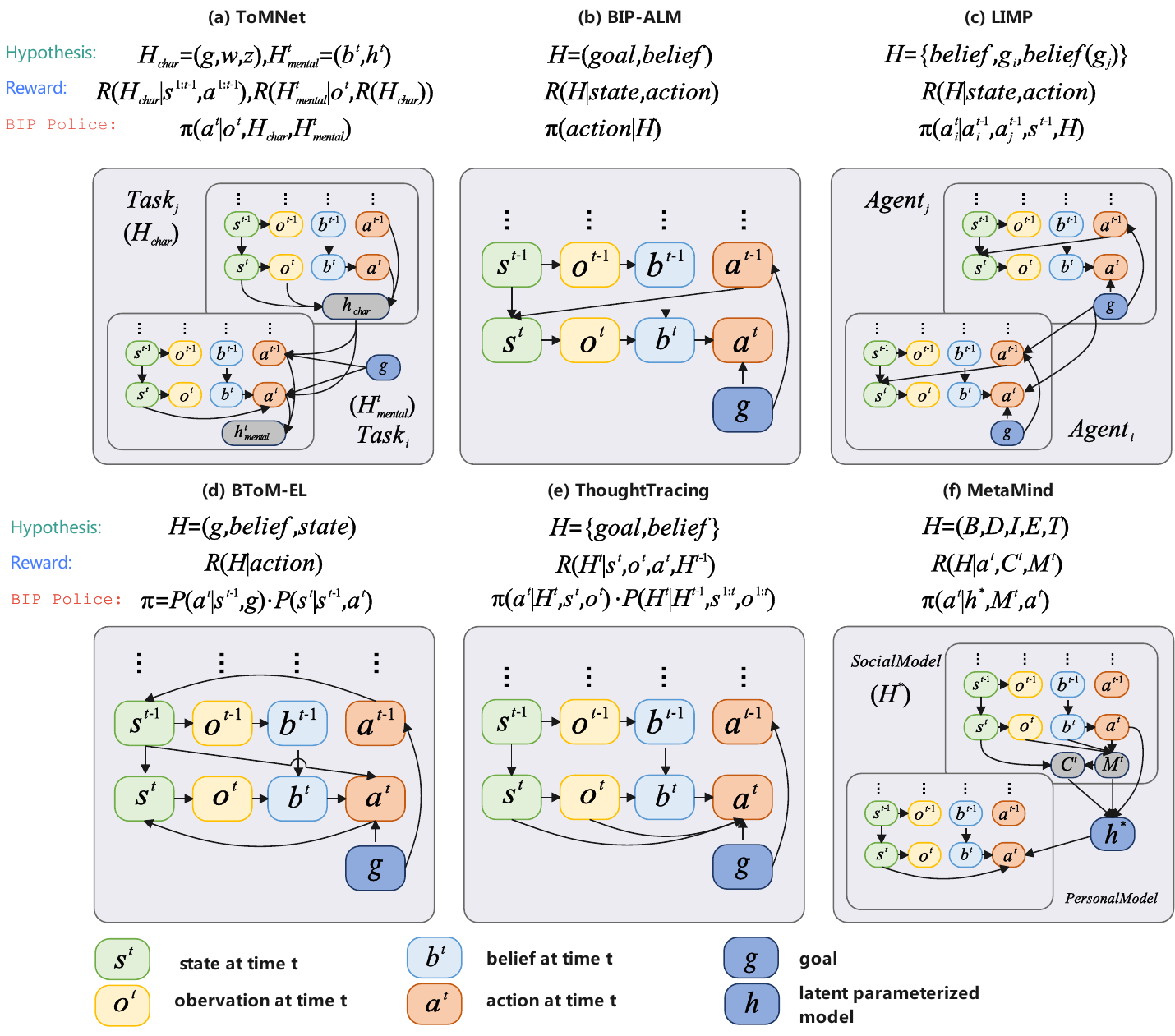}
\caption{A Three-Stage Probabilistic Model Decomposition of Typical Model-Based Inference Methods from the Perspective of Bayesian Theory of Mind.
}\label{fig_prob_graph}
\end{figure}

In this subsection, we perform a unified abstraction and generalization of typical model-based inference methods, yielding the probabilistic graph illustrated in Figure \ref{fig_prob_graph}. The notations of certain methods are standardized here—for example, both actions \(a^t\) and utterances \(u^t\) in LIMP \cite{shi2025muma} are uniformly defined as actions. As shown in Figure \ref{fig_prob_graph}, we decompose these models into three components, namely Assumption, Feedback, and Strategy, whose definitions and functions are described as follows:
\begin{itemize}
\item \textbf{Assumption} refers to the hypotheses about the composition of mental states. The previous chapter analyzed the hypotheses of different psychological and cognitive science schools regarding the constituent elements of the mental world (Table \ref{table_mental_world_element}), while most models and datasets usually adopt simplified representations (Table \ref{table_mental_world_element_datasets}). These representations generally include environmental states \(s^t\), agent actions \(a^t\) (including utterances \(u^t\)), observations \(o^t\), and so on.
\item \textbf{Feedback} essentially optimizes the strategy model by judging the error between the mental states output by the inverse planning model and the ground-truth mental states. The inverse planning model specifies which variables are used to infer the hypothesized mental states. In Bayesian inverse planning methods \cite{baker2017rational,baker2009action,shum2019theory}, the inverse planning model is derived from the strategy model.
\item \textbf{Strategy} denotes the forward planning model that derives the probability distribution of a certain action from the hypothesized mental states. Traditional decision-making models usually need to be acquired via reinforcement learning. By contrast, BIP-ALM \cite{jin2024mmtom} draws on methods such as \cite{huang2022language,li2022pre} and leverages LLMs for decision-making.
\end{itemize}


Inverse planning serves as the core method for inferring goals and intentions within the Model-Based paradigm. Its fundamental assumption is that human behavior constitutes optimal planning executed to achieve specific goals. Thus, intention inference can be realized through the inverse process of "observing behaviors $\rightarrow$ backtracking optimal plans $\rightarrow$ determining goals" \cite{rabinowitz2018machine,baker2009action}.

ToMnet \cite{rabinowitz2018machine} achieves machine Theory of Mind reasoning capabilities via a meta-learning paradigm, constructing a first-order ToM system using a neural network-based implicit reasoning approach. Its core innovation lies in reformulating the acquisition of ToM capabilities as a meta-learning problem: by observing the behavioral trajectories of numerous agents, ToMnet autonomously learns to model others’ mental states, rather than relying on manually designed agent models (e.g., traditional inverse reinforcement learning or Bayesian ToM methods). Formally, ToMnet does not directly establish a Bayesian inference model. However, its built character net \(H_{char}\) and mental state net \(H_{ms}\) implicitly model the relationships among state variables of various mental world models, as illustrated in Figure \ref{fig_prob_graph}(a). Architecturally, ToMnet adopts a three-module collaborative design:
\begin{itemize}
\item The character net extracts cross-episode agent traits to form priors.
\item The mental state ne captures transient mental states of the current episode to form posteriors.
\item The prediction net integrates the two to enable behavior prediction.
\end{itemize}
This system demonstrates multi-level ToM capabilities in gridworld environments: it can infer the goal-directed behaviors of agents, understand false beliefs arising from partial observability (passing the Sally-Anne test), and even identify unannotated behavioral patterns among trained agents (e.g., left-hand/right-hand wall-following strategies). Although ToMnet is an end-to-end black-box model, variational information bottleneck technology enables partial disentanglement of the embedding space, revealing the distribution of agents across trait dimensions. This work pioneered the proof that ToM reasoning can emerge from purely data-driven neural learning, providing a new technical pathway for multi-agent interaction, human-machine collaboration, and AI interpretability. Nevertheless, the implicitness of its reasoning process and the lack of structured symbolic reasoning remain key directions for future research to address.
BIP-ALM \cite{jin2024mmtom} integrates inverse planning, Bayesian reasoning, and language models to form a multimodal intention inference framework, as shown in Figure \ref{fig_prob_graph}(b). The framework proceeds in three steps:
\begin{itemize}
    \item First, it parses users’ natural language descriptions (e.g., "I want to drink water") via a language model to obtain preliminary goal clues.
    \item Second, it observes users’ action sequences (e.g., "reaching toward the cabinet") through visual sensors.
    \item Finally, based on Bayesian inverse planning, it calculates the matching probability between different candidate goals (e.g., "grabbing a cup", "grabbing a bottle") and the observed language-action information, and selects the goal with the highest probability as the inference result.
\end{itemize}
The innovation of BIP-ALM lies in fusing multimodal information via Bayesian reasoning, quantifying the uncertainty of different clues (e.g., weighting action clues more heavily when language descriptions are ambiguous), thereby enhancing the robustness of goal inference.

To adapt to multi-agent interaction scenarios, researchers have extended the Model-Based paradigm to the joint reasoning of multi-agent mental states, with LIMP \cite{shi2025muma} emerging as a representative approach. LIMP constructs a three-layer reasoning architecture:
\begin{itemize}
    \item First, the multimodal information fusion module leverages Gemini 1.5 Pro to extract action sequences from videos, and employs GPT-4o to parse textual dialogues, fill gaps in visual perception, and reconstruct the initial environmental state.
    \item Second, the hypothesis parsing module generates hypothesis combinations regarding agents’ beliefs, social goals (help/hinder/independent), and their beliefs about others’ goals for each question option.
    \item Third, the inverse multi-agent planning module, built on the Interactive POMDP (I-POMDP) framework, performs Bayesian reasoning to calculate the posterior probability of each hypothesis. This is achieved by evaluating the likelihood of actions and utterances at each time step under the given hypotheses, with the strategy estimation capability of GPT-4o. The probabilistic dependency graph of its mental state variables is illustrated in Figure \ref{fig_prob_graph}(b).
\end{itemize}
Experimental results show that LIMP achieves an accuracy of 76.6\% on the MuMA-ToM benchmark, significantly outperforming baseline models such as GPT-4o (50.6\%) and Gemini 1.5 Pro (56.4\%). It demonstrates particularly outstanding performance on complex tasks including social goal reasoning (67.7\%) and goal belief reasoning (68.7\%).

BToM-EL \cite{ying2024grounding} proposes a belief language understanding framework based on first-order ToM reasoning, which anchors natural language belief statements in Bayesian Theory of Mind via a neuro-symbolic fusion paradigm. The core innovation of this method lies in providing functional role semantics for beliefs, which proceeds in three key steps:
\begin{itemize}
    \item First, it leverages large language models to automatically convert natural language belief statements into symbolic expressions of epistemic logic.
    \item Second, it performs reasoning on observed agent behaviors through Bayesian inverse planning, jointly inferring a consistent distribution of goals, beliefs, and plans that explain the behaviors.
    \item Third, it uses epistemic logic to evaluate the truth value of belief statements against the inferred belief states. The dependency of its mental state variables is illustrated in Figure \ref{fig_prob_graph}(d).
\end{itemize}
This hybrid reasoning architecture fully exploits the complementary advantages of probabilistic reasoning and symbolic reasoning: Bayesian reasoning captures the uncertainty and gradedness of inferences, while symbolic logic ensures the compositionality of belief statements. Experimental results show that the model significantly outperforms non-mentalizing baselines (\(\rho=0.19\)) and heuristic mentalizing models (\(\rho=0.04\)) in predicting human goal attributions for agents and belief statement ratings, achieving a correlation of \(\rho=0.92\) with human judgments. This validates the critical role of ToM reasoning in understanding belief language. The method also features high interpretability—all reasoning steps, ranging from natural language parsing and probability distribution sampling to logical evaluation, are traceable and verifiable.

Inspired by the Bayesian Theory of Mind framework and drawing on the architecture of the Sequential Monte Carlo algorithm, Thought-tracing \cite{kim2025hypothesis} transforms the ToM reasoning problem into a process of probabilistic inference regarding agents’ mental states. The dependency of its mental state variables is illustrated in Figure \ref{fig_prob_graph}(e). The core mechanism of the algorithm consists of three steps:
\begin{itemize}
    \item Generating multiple natural language hypotheses about the target agent’s beliefs and intentions at each time step.
    \item Updating the weights of these hypotheses based on the likelihood of the agent’s actions, prioritizing the retention of more reasonable interpretations.
    \item Maintaining hypothesis diversity through resampling and rejuvenation to avoid particle degeneracy.
\end{itemize}
This algorithm supports both first-order and second-order ToM reasoning. It significantly improves the performance of various large language models (LLMs) such as GPT-4o and Llama 3.3 across four benchmark tests, and even outperforms specialized reasoning models (e.g., o3-mini, R1). Thought-tracing features strong interpretability: it outputs structured reasoning trajectories, explicitly labeling the agent’s perceptions, actions, and belief states at each moment to form a complete evolutionary chain of mental states. Experiments further reveal a significant distinction between social reasoning and mathematical/programming reasoning: reasoning models do not exhibit consistent advantages in ToM tasks and face a performance trade-off between true-belief and false-belief scenarios, indicating that the field of social reasoning requires independent training and reasoning strategies.

MetaMind \cite{zhang2025metamind} addresses the deficiency of large language models (LLMs) in Theory of Mind (ToM) capabilities for social reasoning, proposing a three-stage collaborative reasoning framework inspired by the metacognitive theory in psychology. This method decomposes social understanding into a cognitive chain, with the dependency of its mental state variables illustrated in Figure \ref{fig_prob_graph}(f). The framework proceeds as follows:
First, the Theory-of-Mind Agent generates multi-dimensional mental state hypotheses (covering five dimensions: beliefs, desires, intentions, emotions, and thoughts) based on dialogue context and social memory, supporting 1st–2nd order recursive mental reasoning.
Second, the Moral Agent optimizes and filters these hypotheses according to cultural norms, ethical constraints, and role expectations, adopting a weighted scoring mechanism based on information gain and situational rationality (Equation 1).
Finally, the Response Agent generates responses by integrating the optimal hypotheses with the user’s long-term social memory, and triggers iterative revision via empathy-coherence dual evaluation (Equation 2).
This framework adopts a neuro-symbolic fusion paradigm. While retaining the semantic generation capabilities of LLMs, it introduces symbolic mental state annotation and probabilistic constrained reasoning, achieving traceable reasoning chain outputs (including hypothesis types, evidence bases, and constraint violation detection). Experimental results on three benchmarks including ToMBench demonstrate that MetaMind improves the average ToM reasoning accuracy of 16 mainstream LLMs (e.g., GPT-4, Claude-3.5, DeepSeek-R1) by 6.2\%, with a 35.7\% performance boost in real-world social scenarios. For the first time, it enables models to reach human-level performance on key ToM tasks, verifying the effectiveness of structured metacognitive decomposition for enhancing AI social intelligence.

AutoToM \cite{zhang2025autotom} further enables the automated construction of mental models, with its core being an Agent Model Discovery mechanism: instead of manually defining the types of mental states (e.g., beliefs, goals), it automatically learns the representational dimensions of mental states (e.g., "cognition of object positions", "judgment of task priorities") from large-scale interaction data via Bayesian nonparametric models, and constructs corresponding reasoning rules. For instance, during interactions with different users, AutoToM can automatically identify that "User A pays more attention to the accuracy of object positions, while User B prioritizes task completion speed", and adjust the parameters of the mental model for different users accordingly. The advantages of AutoToM are twofold: it reduces the cost of manual design and enhances the model’s adaptability to different users.
Nevertheless, the limitations of the Model-Based paradigm are also quite prominent:
Limited generalization of symbolic representations: Methods such as BIP-ALM \cite{jin2024mmtom} still require manual definition of certain symbols (e.g., "goal types"). When confronted with undefined novel mental states (e.g., complex emotions such as "jealousy" and "suspicion"), the model is unable to process them.
High computational complexity: Bayesian reasoning and inverse planning require traversing a large number of possible mental state hypotheses, resulting in poor real-time performance, which makes it difficult to meet the real-time interaction requirements of embodied agents.

The Model-Based Inference paradigm, centered on the principle of "explicitly constructing mental models", systematically addresses the "black-box" issue of the Prompting paradigm—where performance relies on the implicit capabilities of LLMs—via mathematical and algorithmic frameworks such as Bayesian inverse planning and neuro-symbolic fusion. It has formed a technological evolution path that expands from single-agent to multi-agent scenarios, from goal inference to belief language understanding, and from manual design to automated modeling.
From the meta-learning data-driven approach of ToMnet \cite{rabinowitz2018machine}, to the multimodal fusion of BIP-ALM \cite{jin2024mmtom}; from the multi-agent joint reasoning of LIMP \cite{shi2025muma} and belief language adaptation of BToM-EL \cite{ying2024grounding}, to the probabilistic trajectory modeling of Thought-tracing \cite{kim2025hypothesis} and social reasoning optimization of MetaMind \cite{zhang2025metamind}, and finally to the automated model construction of AutoToM \cite{zhang2025autotom}, methods under this paradigm have gradually enhanced the interpretability, controllability, and scenario adaptability of ToM reasoning. Certain approaches (e.g., MetaMind \cite{zhang2025metamind}, LIMP \cite{shi2025muma}) have demonstrated ToM capabilities that are comparable to or even exceed human performance in both benchmark tests and real-world scenarios.
However, two critical obstacles still impede its widespread deployment: the generalization bottleneck of symbolic representations (which requires manual definition of core symbols and struggles to handle complex emotion-based mental states) and high computational complexity (resulting in insufficient real-time performance). Future research needs to make further breakthroughs in directions such as automated symbol generation and lightweight probabilistic reasoning, to better adapt to application scenarios with higher requirements for real-time performance and scenario diversity, such as embodied agents.

\section{Evolution of ToM Evaluation Benchmarks}

\begin{figure}[!htb]
\centering
\includegraphics[width=1.0\textwidth]{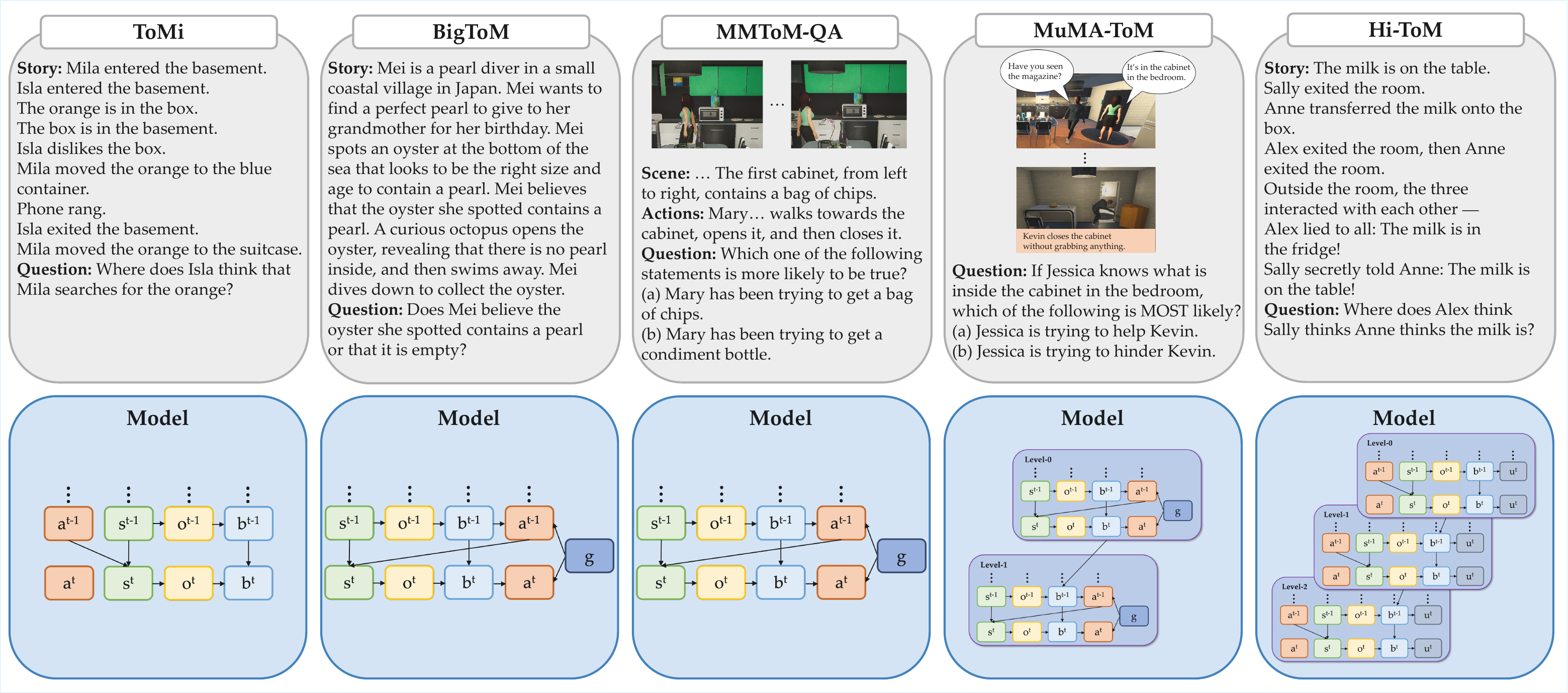}
\caption{Typical ToM datasets with corresponding agent models tailored by AutoToM \cite{zhang2025autotom}. 
}\label{fig_tom_data}
\end{figure}

The fundamental objective of evaluating physical world models is to construct specific video clips that eliminate the models’ reliance on scene continuity, while probing their performance in aspects such as physical persistence and physical causality. Representative benchmarks include Minimal Video Pairs \cite{krojer2025shortcut}, IntPhys2 \cite{bordes2025intphys}, CasualVQA \cite{foss2025causalvqa}, and WorldPrediction \cite{chen2025worldprediction}.
In contrast, the evaluation of mental world models primarily focuses on assessing mental reasoning capabilities. As illustrated in Table \ref{table_ToM_data}, such evaluations have evolved from early text-based benchmarks to multimodal benchmarks, and further toward dynamically interactive benchmarks.
\begin{table}[!htb]
    \centering
    \setlength{\belowcaptionskip}{0.2cm}
    \caption{ToM Reasoning Benchmarks. Exp. means Expandable. Int. means interactive.}
    \scalebox{0.72}{
    \begin{tabular}{llllp{2cm}p{6cm}p{6.5cm}}
    \toprule
   \textbf{Dataset}	& \textbf{Year}&   \textbf{Exp.}& \textbf{Int.}&\textbf{Modality}	&\textbf{Task Target/Application Scene} 	&\textbf{Dataset Detail} \\
   \midrule
   ToMi \cite{le2019revisiting}	& 2019&\checkmark& \ding{53} 	&Text& Synthetic narrative scenarios through question answering & six distinct question types: Reality, Memory, 1st and 2rd order belief of two agents\\
BigToM \cite{gandhi2023understanding}	& 2023 &\checkmark &\ding{53}	 & Text & fictional stories for evaluating "percepts to beliefs", "percepts to actions", and "actions to beliefs"& 5,000 evaluations generated from 200 causal templates focusing on 6 conditions
\\
 CoToMA \cite{li2023theory} &2023&\checkmark&\checkmark & Text & multi-agent collaborative scenario in rescue missions& 3 question types (introspection, 1st-/2rd- order ToMs) \\
FANToM \cite{kim2023fantom}&2023&\checkmark&\ding{53} &Text & information-asymmetric multiparty conversation scenarios & 256 conversations with 10K questions across 6 types \\

Hi-ToM	\cite{wu2023hi} &2023& \checkmark&\ding{53}	& Text& High-order ToM evaluation in fictional Sally-Anne-like stories & Average 26.47 lines with 5 agents/questions per story\\
High-order ToM \cite{street2024llms} & 2024 & \ding{53} &\ding{53} &Text& High-order ToM evaluation in fictional social interaction scenarios & 7 stories, 140 statements, orders 2-6\\
ExploreToM \cite{sclar2024explore}	&2024 & \checkmark&\ding{53}&Text&  fictional narrative scenarios&	 1,620 stories, 1st/2rd order and state-tracking questions
	\\
 MAgIC \cite{xu2024magic} &2024& \ding{53}& \checkmark&Text & ToM evaluation in multi-agent competitions & 103 competition cases across 5 scenarios with 7 metrics  \\
 TestingToM \cite{strachan2024testing} & 2024 & \ding{53}& \ding{53}&Text & Social cognition evaluation &ToM testing with novel items and variants    \\
EAI \cite{mozikov2024eai} & 2024 &\checkmark&\checkmark&Text& ToM evaluation via existing ethical and game-theoretic scenarios &ETHICS, MoralChoice, and StereoSet subsets \\
 COKE \cite{wu2024coke} & 2024& \checkmark &\ding{53}&Text&  ToM evaluation in daily social situations across five topics&  45,369 cognitive chains, 62,328 nodes, 1,200 situations, 5 topics, 4  tasks\\
 OpenToM \cite{xu2024opentom} & 2024& \checkmark &\ding{53}&Text& ToM evaluation in fictional stories with personified characters & 696 narratives, 13,708 questions covering location tracking, multi-hop reasoning, and attitude inference\\
 LLM-ToM \cite{kosinski2024evaluating} & 2024 &\ding{53}&\ding{53} & Text & False-belief understanding in fictional narrative scenarios & 40 tasks, 2 types, 8 scenarios and 16 prompts per task \\
 DynToM \cite{xiao2025towards} & 2025& \checkmark &\ding{53}&Text&  ToM evaluation in social interaction scenarios  & 1,100 contexts, 5,500 scenarios, 78,100 questions, 4 question types\\

Decrypto \cite{lupu2025decrypto} &2025& \checkmark  &\checkmark& Text& mental states reasoning while encrypting/decrypting messages & 10 human games, up to 8 turns per game \\
 SocialMaze \cite{xu2025socialmaze}&2025 & \checkmark &\ding{53}&Text& Social deduction games, daily life interactions, and digital community&  70,000 total instances across 6 tasks in 3 scenarios\\
UniToMBench \cite{thiyagarajan2025unitombench} &2025& \ding{53} &\ding{53}&Text& fictional story scenarios and social interaction contexts & 1,025 hand-written scenarios with 8 TOMBENCH task categories\\
 XToM \cite{chan2025xtom} & 2025&\ding{53}&\ding{53}&Text& Multilingual ToM evaluation across fictional stories & 300 stories/dialogues, 5 languages, 3 sub-tasks \\
\midrule
Watch-And-Help \cite{puig2020watch} & 2020 & \checkmark &\checkmark&\multirow{1}{2cm}
{Graphs, Video, etc.}& Social perception and collaboration evaluation in household scenarios &  1011 training tasks, 5 activity categories, 30 predicate type\\
NOPA \cite{puig2023nopa}& 2023 & \checkmark &\ding{53} & Visual 3D & Embodied household multi-agent collaboration scenarios &  10 testing episodes, 40 human trials, household tasks in virtual homes \\
MMToM-QA \cite{jin2024mmtom} &  2024 &\checkmark &\ding{53} &Video+Text &  ToM evaluation in household activity scenarios & 134 videos, 600 questions, 7 question types, \~1,462 frames per video	\\
MuMAToM \cite{shi2025muma}	& 2025& \checkmark	&\ding{53} &Video+Text& multi-agent embodied household collaboration scenarios & 4 apartment environments with 900 questions of 3 types	\\
ChARTOM \cite{bharti2025chartom} &2025&\ding{53}&\ding{53} &Image+Text&  ToM evaluation in misleading visual interpretation scenarios &   15 manipulation groups, 30 charts in pairs, 5 chart types, 2 question types \\
SoMi-ToM \cite{fan2025somi} &2025& \checkmark& \checkmark& Video, Image, Text&Embodied multi-agent collaborative crafting scenarios & 35 videos, 363 images, 1225 questions across 3 inference tasks  \\
ToMLoc \cite{chen2025through} & 2025&\ding{53}&\ding{53}&Video+Text& ToM evaluation in real-world social interaction scenarios & 1,403 videos, 8,076 questions, 4-choice format\\
GridToM \cite{li2025black} & 2025 & \checkmark&\ding{53}&Video+Text& ToM evaluation in 2D grid world multi-agent scenarios & 1,296 samples, 27 map layouts, 3 question types\\

\bottomrule
    \end{tabular}}
    \label{table_ToM_data}
\end{table}

\subsection{Early Benchmarks: Laying Foundations and Exposing Limitations}

As research into the social intelligence of embodied agents advances, early researchers transformed classic ToM tasks from psychology (e.g., false-belief tasks, goal-inference tasks) into quantifiable Machine Reading Comprehension (MRC) or Question Answering (QA) tasks, establishing the first batch of ToM evaluation benchmarks. Among these, ToMi \cite{le2019revisiting} stands out as the most representative, as illustrated in Figure \ref{fig_tom_data}.
Targeting the core flaw of the early ToM-bAbi benchmark—namely, that its data contained exploitable fixed patterns (e.g., stories followed rigid templates, and problems could be solved perfectly via heuristic rules such as "locating the first/last position of an object" or "checking for the presence of the word 'exited'", without requiring genuine ToM reasoning)—ToMi \cite{le2019revisiting} proposed a systematic improvement scheme with three key components:
\begin{itemize}
    \item It adopted a unified random generation mechanism to construct stories involving true beliefs, false beliefs, and second-order false beliefs, thereby eliminating idiosyncratic biases across different story types.
    \item It introduced interference elements such as irrelevant agent actions, distracting sentences about locations, and randomized action sequences to reduce data predictability.
    \item It mandated the generation of a full set of question types for each story, including Reality, Memory, first-order belief (e.g., "Where will Agent A look for the object?"), and second-order belief (e.g., "Where does Agent A think Agent B will look for the object?"). Moreover, it innovatively proposed an aggregate accuracy evaluation metric: a story was deemed successfully reasoned only if all its associated questions were answered correctly. This metric ensured that models truly distinguish between the objective state of the world and the subjective mental states of agents.
\end{itemize}
Nevertheless, ToMi \cite{le2019revisiting} still has notable limitations. Its evaluation data consists of fully synthetic simple stories with a single linguistic scenario and low complexity, failing to address key challenges in real natural language such as coreference resolution, semantic ambiguity, and commonsense association. Furthermore, its reasoning tasks focus solely on the single dimension of "object-location beliefs", and do not cover complex ToM requirements essential for embodied agents in real interactions, such as "emotional mentalizing", "dynamic goal inference", and "multi-agent collaborative game theory". Thus, ToMi \cite{le2019revisiting} can only serve as a basic entry-level test for machine ToM capabilities, rather than an evaluation standard for comprehensive social intelligence.

A mental world model is not a static "mental snapshot", but rather a "living cognitive system" that needs to be dynamically updated amid continuous social interactions. The core advantage of human ToM capabilities lies in the ability to revise judgments about others’ mental states in real time based on new interactive information (e.g., others’ new behaviors, new environmental changes). For instance, if a user originally holds the false belief that "the cup is in the drawer", but finds the drawer empty upon opening it, humans will immediately infer that the user’s belief has been updated to "the cup is not in the drawer", and that the user may experience the emotion of "confusion". If an agent’s mental model lacks the capacity for dynamic updating, it will still provide assistance based on the outdated assumption that "the user thinks the cup is in the drawer", resulting in disjointed interactions.
ExploreToM \cite{sclar2024explore} highlights the importance of dynamic updating by designing "adversarial scenarios". For example, in one such scenario: Agent A hides an object at location X and then leaves; Agent B first moves the object to Y, then moves it back to X (creating a "double transfer"). In this case, the agent is required to track the evolution of A’s beliefs: A initially believes the object is at X → if A does not observe B’s movements, A will still believe the object is at X → if A observes B’s first transfer (to Y) but not the second (back to X), A will believe the object is at Y. This benchmark generates adversarial data based on a domain-specific language using the A* search algorithm.
Employing a text modality, the dataset focuses on evaluating mental reasoning capabilities in fictional narrative scenarios, comprising 1,620 story structures that cover 162 configuration combinations. Unlike traditional interactive environments, ExploreToM adopts an evaluation method that pairs static stories with automatically generated questions, with question types spanning multiple levels including first-order belief, second-order belief, and state tracking. The core innovation of this framework lies in two aspects: ensuring annotation reliability via a precise mental state tracker, and leveraging an adversarial search strategy to targetedly generate hard samples that challenge state-of-the-art models.
Experimental results show that this dataset successfully exposes significant flaws in ToM reasoning among advanced models such as GPT-4o and Llama-3.1, with accuracy rates as low as 0–9\%—fully demonstrating its value as a robustness evaluation benchmark. Furthermore, models fine-tuned on this dataset achieve a 27-percentage-point performance improvement on the classic ToMi \cite{le2019revisiting} benchmark, which verifies its effectiveness as training data.

BigToM \cite{gandhi2023understanding} achieves favorable scalability via programmatic generation, leveraging GPT-4 to populate 200 elaborately designed causal templates and ultimately producing a large-scale evaluation corpus containing 5,000 test samples. Proceeding from the core competencies of mental reasoning, BigToM systematically examines three key dimensions: forward belief reasoning (inferring others’ beliefs from perceptual cues), forward action reasoning (predicting others’ behaviors from perceptual information), and the most challenging backward belief reasoning (retrospectively deducing others’ belief states from observed behaviors). The dataset incorporates 25 refined control conditions, which enable the differentiation between true-belief and false-belief scenarios and effectively isolate and diagnose the failure modes of models across distinct reasoning stages. All scenarios within the dataset are fictional daily-life situations involving diverse roles such as baristas, bakers, fishermen and teachers, thus ensuring content richness and ecological validity. Nevertheless, BigToM currently adopts a static text-based story-question-answer paradigm and has not yet implemented an interactive evaluation environment.

Recursive reasoning is a core characteristic of ToM operation, referring to the cognitive agent’s ability to represent "one individual’s cognition of another individual’s mental state". It can be categorized into first-order ToM ("I think you believe X"), second-order ToM ("I think you think he believes X") and high-order ToM (third-order and above, e.g., "A thinks B thinks C thinks D believes X"). In simple social interactions, first-order ToM is sufficient to meet needs (e.g., a robot infers "the user thinks the cup is in the drawer"), whereas in complex scenarios (such as multi-person collaboration, negotiation, and deception), high-order ToM is a prerequisite for effective interaction.
To address the inadequacy of scenario complexity in early benchmarks, researchers have begun to explore "high-order ToM" evaluation, with Hi-ToM \cite{wu2023hi} (High-Order Theory of Mind) emerging as a typical representative. The core innovation of Hi-ToM lies in expanding the depth of ToM reasoning from first-order to third-order and above, and introducing complex social dynamic scenarios such as deception and separation of public and private information. For instance, Hi-ToM designs a "secret transmission" scenario: "Alice, Bob and Charlie are together; Alice privately tells Bob ‘the box contains candies’ (it actually contains toys), and Bob knows that Charlie did not hear this sentence. Later, Charlie asks Bob ‘what is in the box’, and Bob says to Charlie ‘it is a toy’". The high-order QA pairs for this scenario are: "What does Alice think Bob will tell Charlie is in the box?" (second-order reasoning, correct answer: "candies"); "What does Bob think Alice thinks Bob will tell Charlie is in the box?" (third-order reasoning, correct answer: "candies"). In addition, Hi-ToM also introduces "deceptive communication" scenarios (e.g., a subject deliberately transmits false information to mislead others) to test the model’s ability to understand "inconsistency between intention and linguistic content". Compared with ToMi \cite{le2019revisiting}, the advancements of Hi-ToM are as follows:
\begin{itemize}
    \item It increases reasoning depth and exposes the shortcomings of models in high-order recursive reasoning.
    \item It incorporates complex social dynamics, making it more closely aligned with real interpersonal interaction scenarios.
\end{itemize}
High-order ToM \cite{street2024llms} is another text dataset specifically designed for evaluating high-order theory of mind reasoning capabilities. This dataset focuses on assessing large language models’ ability to perform 2nd-to-6th-order recursive mental state reasoning in fictional social interaction story scenarios (e.g., "I think you believe she knows"), which constitutes a core component of human social intelligence. Adopting a static question-answer format, the dataset comprises 7 short stories depicting multi-person social interactions, with each story involving 3–5 characters and accompanied by 140 true-false questions. It provides a rigorous test benchmark for comparing the performance of models and humans in high-order mental reasoning tasks, and holds significant implications for understanding the reasoning capabilities of LLMs in complex multi-party social scenarios.

Nevertheless, both Hi-ToM \cite{wu2023hi} and High-order ToM \cite{street2024llms} have notable limitations: first, their modality is still confined to text, failing to evaluate the model’s ability to infer mental states from multimodal signals such as vision and actions, which constitutes a core requirement for embodied agents. Second, the scenarios still adopt a static "story reading + offline QA" paradigm, where the model does not need to update mental states in real time during interaction and only has to perform one-time reasoning on fixed stories, which is inconsistent with the practical demand of embodied agents for "reasoning while interacting". Finally, data generation still relies on manually designed story templates—though more flexible than ToMi \cite{le2019revisiting}, these templates still have certain predictable patterns, and models may achieve high scores by learning heuristics such as "answers to high-order questions tend to correspond to initial information".

\subsection{Paradigm Shift: Multimodality and Dynamic Interaction}

In 2025, Social World Models, proposed by Zhou et al. \cite{zhou2025social}, pointed out that AI systems face fundamental challenges in understanding social dynamics: traditional methods mainly rely on static text to learn social interactions, yet such text suffers from reporting bias, lacks explicit expressions of mental states, and often adopts an omniscient perspective. By contrast, ToM reasoning in real human social interactions depends on the fusion of multimodal cues such as language, vision, and actions, rather than unimodal information. For instance, humans can quickly infer that the other party is in an "anxious" state and holds the belief that "the item is lost" through multimodal signals including frowning (visual) + elevated tone of voice (auditory) + "This thing is missing" (linguistic). If only a single modality is available (e.g., just hearing the linguistic content), it may be mistakenly judged as a "calm inquiry". This characteristic requires the ToM mechanism of embodied agents to be equipped with multimodal integration capabilities, and the emerging multimodal ToM benchmarks in recent years have been designed precisely based on this demand.

The early Watch-And-Help \cite{puig2020watch} benchmark supports four modalities simultaneously, namely symbolic graphs, RGB videos, depth maps, and semantic maps. Its core innovation lies in decomposing mental reasoning evaluation into two phases: observation-reasoning and collaboration-verification. Specifically, the agent first needs to infer the implicit goals of human partners by watching demonstration videos (social perception capability), and then collaborates with humans in new environments to efficiently accomplish these goals (collaborative planning capability).
The dataset consists of 1011 training tasks covering five categories of household activity scenarios: table setting, food storage, meal preparation, dishwashing, and leisure reading. These tasks are generated by combining 30 predicate types, enabling programmatic scalability. The evaluation environment provides multimodal observations including symbolic graphs, RGB videos, depth maps, and semantic maps, and incorporates a planning-based human simulation agent with bounded rationality that can respond in real time to behavioral changes of collaborative partners. The unique feature of this dataset lies in its interactive design: the agent’s mental reasoning capability is measured not only by static goal recognition accuracy, but also comprehensively by acceleration ratio and success rate in dynamic collaboration. This requires the agent not only to understand others’ intentions, but also to predict their future actions, proactively update their belief states, and handle multi-level action planning. This evaluation paradigm of "verifying understanding through collaboration" makes it an important benchmark for testing practical mental reasoning capabilities in embodied scenarios.
The NOPA dataset \cite{puig2023nopa} focuses on multi-agent collaborative tasks in household scenarios, requiring assistive robots to perform online goal inference while observing human behaviors—a typical application of Theory of Mind (ToM). Built on a photorealistic 3D virtual household environment, the dataset comprises 10 test scenarios and 40 human interaction trials, covering daily household chores such as table setting and item organization. Unlike traditional static mental reasoning evaluations, this dataset emphasizes dynamicity and uncertainty handling: the agent needs to continuously update its inference of human intentions when goal information is ambiguous, and adjust collaborative strategies accordingly. This design allows the dataset to evaluate not only the accuracy of goal recognition, but also the depth of the agent’s understanding of others’ mental states in real collaboration, the timeliness of reasoning, and decision-making capabilities based on uncertainty, thereby providing an important evaluation platform for research on social intelligence and mental reasoning in embodied environments.
Despite offering rich multimodal observation dimensions, datasets of this kind have not become the mainstream of current multimodal ToM evaluation due to constraints related to data acquisition costs and processing resource consumption.

As the application scenarios of embodied agents expand toward multimodal interaction, low-cost multimodal ToM benchmarks combining video and text have emerged as the times require, enabling a paradigm shift from "text-based reasoning" to "multimodal joint reasoning". The core feature of such benchmarks lies in the introduction of visual modal inputs such as videos and images, requiring models to fuse linguistic and visual information for mental state inference—this is far more aligned with the practical working mode of embodied agents, which "observe behaviors via cameras + understand demands through voice interaction".
The MMToM-QA \cite{jin2024mmtom} dataset is a representative video-text multimodal Theory of Mind evaluation benchmark designed to test machines’ capability of reasoning about human mental states. Integrating two modalities (video and text), this dataset simulates the cognitive process through which humans comprehend others’ intentions from multi-source information in real scenarios. Focusing on daily activity scenarios in household environments, it comprises 134 video clips and 600 carefully designed question-answer items, systematically evaluating models’ performance in two core dimensions: goal reasoning and belief reasoning.
The core advantage of this dataset lies in its scalable programmatic generation method: it synthesizes human behavior sequences in simulated environments via a POMDP planner, automatically generating training data with ground-truth annotations. The question types cover seven categories, including true belief, false belief, long-term belief tracking, and goal reasoning based on different belief conditions, comprehensively assessing the depth of models’ understanding of complex mental states. By requiring models to integrate dynamic behavioral information from videos and scene state descriptions from text, this dataset effectively verifies whether models possess human-like multimodal mental reasoning capabilities, providing an important evaluation tool for building AI systems that are more consistent with human social cognition.
Unlike the MMToM-QA \cite{jin2024mmtom} benchmark, which focuses on single-agent belief tracking, MuMA-ToM \cite{shi2025muma} is designed to evaluate how artificial intelligence systems infer mental states in real household interactions involving two agents.
This dataset targets three key dimensions of multi-agent ToM reasoning: belief inference (what agents know about the environment), social goal inference (whether agents intend to help, hinder, or act independently), and goal belief inference (agents’ cognition of each other’s goals). Among the total 900 questions, every single one requires the integration of cross-modal information—agents’ verbal interactions are presented in the form of text/subtitles, while their specific actions and outcomes are depicted in videos. This design aims to test models’ ability to track how conversations update beliefs and how subsequent actions reveal underlying intentions.
This benchmark is generated programmatically using VirtualHome and GOMA, featuring strong scalability while maintaining ecological validity. By systematically sampling agent goals, initial beliefs (true/false/uncertain), and social intentions across four household environments, it creates diverse scenarios to isolate specific ToM reasoning challenges. Human participants achieved an accuracy rate of 93.5\%, establishing a solid performance upper bound, whereas state-of-the-art large multimodal models (LMMs) only reached 50–56\%, revealing a significant human-machine gap in ToM capabilities. Adopting a non-interactive question-answer format, this dataset enables the systematic evaluation of ToM reasoning capabilities while eliminating interference from requirements related to action execution or online planning.

CHARTOM \cite{bharti2025chartom} is the first visual Theory of Mind benchmark dataset designed for misleading charts, aiming to evaluate the capability of large multimodal models (LMMs) to understand and reason about how humans are influenced by misleading data visualizations. Uniquely, this dataset extends mental reasoning tasks from traditional text comprehension to the domain of visual perception, covering five common chart types including line charts, bar charts, scatter plots, maps, and pie charts. It consists of 30 paired charts corresponding to 15 manipulation types (such as truncated Y-axis, inverted Y-axis, 3D effects, etc.). Two categories of questions are designed for each chart: FACT questions test the understanding of objective information presented in the charts, while MIND questions assess the model’s ability to predict the degree to which humans are misled. The core innovation of the dataset lies in calibrating the Human Misleading Index (HMI) as the ground-truth label for Theory of Mind questions through experiments involving 68 human subjects, quantifying the actual misleading effect of each chart on humans. Evaluation results show that current mainstream large language models perform poorly on this dataset—they not only struggle to accurately predict the misleading nature of charts on humans, but also exhibit significant deficiencies in comprehending the objective information within the charts. This highlights substantial research potential for advancing models’ visual Theory of Mind reasoning capabilities.

The GridToM \cite{li2025black} dataset is built on a 2D grid-world environment, comprising 1,296 samples that cover 27 distinct map layouts. Unlike previous studies, the core innovation of GridToM lies in providing multi-perspective perceptual information for each event—it includes not only an omniscient view, but also video and textual descriptions from both the protagonist’s perspective and the participants’ perspectives, thereby enabling precise control and verification of the model’s ability to distinguish between different cognitive perspectives.
Adopting a video-text bimodal design, the dataset systematically evaluates three types of belief reasoning tasks: initial belief comprehension, first-order belief reasoning, and second-order belief reasoning. Each task is further divided into true-belief (TB) and false-belief (FB) scenarios. By simulating agent behaviors such as entering/exiting rooms and opening/closing doors in the grid world, the dataset creates information-asymmetry contexts—when a door is closed, the agent inside the room cannot perceive events occurring outdoors, thus naturally generating false-belief scenarios. This design ensures that the evaluation of a model’s mental reasoning capabilities no longer depends on the quality of perception of complex real-world scenarios, but instead focuses purely on the ability of perspective separation and belief reasoning.
Notably, GridToM adopts an almost fully automated generation pipeline, where key elements such as map design, path planning, and event timing are controlled via scripts. This ensures data consistency and scalability, laying the foundation for the subsequent construction of more complex mental reasoning evaluation benchmarks. As such, GridToM is not merely an evaluation tool, but also a sustainable research platform.

To bridge the gap between existing Theory of Mind (ToM) evaluations and real-world social interactions, SoMi-TOM \cite{fan2025somi} has constructed an innovative multi-perspective embodied mental reasoning evaluation benchmark. Built on the SOMI multi-agent interaction platform within the Minecraft game environment, this dataset generates rich multimodal interaction data by having three AI agents collaborate or obstruct one another to complete crafting tasks. It comprises 35 third-person perspective videos, 363 first-person perspective images, and 1,225 expert-annotated multiple-choice questions, covering three categories of mental reasoning tasks: state reasoning, goal reasoning, and behavior reasoning.
The core innovation of SoMi-TOM lies in its dual-perspective evaluation framework: the first-person evaluation simulates the agent’s subjective real-time experience, requiring the model to perform real-time state reasoning based on limited multimodal inputs (vision, dialogue, action feedback, etc.). The third-person evaluation provides complete task recordings and textual logs to assess the model’s ability to infer others’ goals and behaviors from an objective global observation. This design not only reflects the characteristics of information asymmetry in real social scenarios, but also enables a comprehensive evaluation of the model’s mental reasoning capabilities in dynamic, complex social interactions. Experimental results show that current state-of-the-art large vision-language models exhibit a significant performance gap compared with humans (a 40.1\% gap in first-person evaluation and a 26.4\% gap in third-person evaluation), revealing the models’ deficiencies in resource tracking, system feedback integration, and distinguishing between intentions and behaviors. With high scalability, this dataset lays a foundation for future research on more complex embodied mental reasoning.

ToMLoc \cite{chen2025through} further extends multimodal ToM evaluation to real-world human social scenarios. It consists of 1,403 video clips sourced from real social interactions on YouTube and 8,076 multiple-choice questions, with four candidate answers provided for each question. Unlike traditional text-only mental reasoning tasks, ToMLoc \cite{chen2025through} focuses on spatiotemporal mental state reasoning in a multimodal (video + text) setting, requiring models not only to answer questions about characters’ beliefs, intentions, and emotions, but also to precisely locate the specific video frames that convey key mental cues.
The core innovation of this dataset lies in the introduction of a fine-grained frame localization task, which demands that models identify the exact moments when subtle social signals (e.g., facial expressions, body language) appear. Although the dataset itself comprises static video-question pairs (non-interactive environment), the paper proposes a method based on median frame approximation and GPT-4V semi-supervised annotation to partially enable data expansion, providing a scalable annotation framework for future research. This dataset fills the gap in video-based mental reasoning evaluation and serves as a critical benchmark for building multimodal AI systems with human-like social cognitive capabilities. Nevertheless, the current reliance on high-quality manual annotation still limits its large-scale expansion, and more advanced automated annotation strategies will be required to overcome this bottleneck in future work.

To synthesize the above analysis, the evolution of ToM evaluation benchmarks presents a clear trajectory: advancing from "single-modality, static reasoning, and individual-level mentalizing" to "multimodality, dynamic interaction, and group-level socializing". Specifically, the first phase (early benchmarks), represented by ToMi \cite{le2019revisiting}, laid the foundation for "evaluating ToM via text-based QA" yet was constrained by templated data and low-order reasoning. The second phase (attempts at higher-order reasoning), represented by Hi-ToM \cite{wu2023hi}, increased reasoning depth and the complexity of social scenarios but still failed to break free from the limitations of text and static settings. The third phase (paradigm shift), represented by MMToM-QA \cite{jin2024mmtom}, MuMA-ToM \cite{shi2025muma}, and ToMLoc \cite{chen2025through}, achieved breakthroughs in multimodality and dynamic interaction, thus being more closely aligned with the practical requirements of embodied agents.





\section{Conclusion}
\subsection{Core Challenges and Future Research Directions}

Despite the remarkable progress achieved in the research on mental world models, the transition from "theory" to "practical application for embodied agents" still confronts four core challenges, which stem from the complexity of mental states, the dynamic nature of embodied interaction, and the disconnect between evaluation and real-world practice.

First, the abstractness of mental states and the complexity of dynamic updating—The elements that a mental world model needs to represent, such as beliefs, goals, and emotions, are highly abstract and cannot be fully captured by precise mathematical models or neural representations. For example, a complex belief like "suspicion" (e.g., "the user suspects the cup may be in the drawer but is not certain") is characterized by fuzziness and uncertainty. Existing models mostly adopt a "binary belief" framework (e.g., "believes it is in the drawer / does not believe it is in the drawer"), and the representation of "uncertainty" is partially approximated by using the token probabilities output by LLMs in methods such as LIMP \cite{shi2025muma}. In contrast, emotional states (e.g., "anxiety", "pleasure") exhibit faster dynamic changes and are influenced by multiple factors including physiology and environment, making it difficult for models to track their intensity and evolutionary trends in real time. Moreover, the updating of mental states is not an isolated event—changes in beliefs may trigger emotional shifts (e.g., the updated belief that "the cup is not in the drawer" leads to the emotion of "confusion"), and emotional changes may in turn affect goal adjustment (e.g., "anxiety" raises the priority of the goal of "drinking water"). This coupled relationship between "beliefs-emotions-goals" causes the complexity of dynamic updating to grow exponentially. Most existing models focus on updating a single mental element (e.g., only updating beliefs) and lack the modeling of coupling relationships among elements, leading to the frequent occurrence of "mental state decoupling" in real-world interactions.

Second, the effective alignment and fusion of multimodal information—Embodied agents perceive the environment and human behaviors through multimodal devices such as visual, auditory, and tactile sensors, yet the information from different modalities exhibits heterogeneity and asynchrony, making effective alignment difficult to achieve. For instance, there may be a time lag (e.g., the action precedes the language by 0.5 seconds) between a visual signal (such as a user reaching out their hand) and a linguistic signal (such as "I want to get the cup"), and the model needs to determine whether the two signals correspond to the same mental state (e.g., the intention of "getting the cup"). If the linguistic signal conflicts with the visual signal (e.g., the user says "I want to get the cup" but reaches for a mobile phone), the model needs to judge which modality is more reliable (e.g., the user may have made a slip of the tongue, and the action should be prioritized). Most existing multimodal ToM models adopt an early fusion strategy (e.g., concatenating visual features and textual features before feeding them into the model), failing to fully account for the uncertainty and priority differences among modalities. In addition, the physical interaction modality in embodied interactions (e.g., physical contact between the robot and the user) has not been effectively integrated—for example, a user conveys the intention of "stopping the action" by "gently pushing the robot’s arm", but existing models struggle to correlate such tactile signals with the corresponding mental state (e.g., "the user does not want the robot to keep getting the cup"), which highlights the limitations of multimodal fusion.

Third, insufficient robustness in high-order recursive reasoning—Embodied agents are required to handle third-order and higher-level ToM reasoning (e.g., "I think you believe that he considers this plan feasible") in complex social scenarios such as multi-person negotiations and team collaboration, yet the robustness of existing models in such scenarios is severely lacking. From a technical perspective, the challenges of high-order reasoning lie in three aspects: (1) Representation explosion—each additional level of reasoning requires the extra storage of hierarchical relationships between "subject-belief-object"; the number of representations to be stored for third-order reasoning is an exponential multiple of that for first-order reasoning, leading to excessive memory consumption of the model. (2) Recursive confusion—models tend to confuse reasoning hierarchies (e.g., misjudging "I think you believe" as "you believe"), and this confusion is further exacerbated by the increase in the number of subjects, especially in multi-agent scenarios. (3) Context dependence—high-order reasoning relies on longer interaction contexts (e.g., "A’s attitude toward B in previous conversations"), but the context windows of existing models are limited (e.g., the context length of most LLMs ranges from $4k$ to $32k$ tokens), making it difficult to handle interactions over a long time span.

Fourth, the disconnect between evaluation and practical application—current benchmarks still face three core challenges: (1) Insufficient data authenticity—the videos in existing multimodal benchmarks are mostly manually recorded or animation-generated, featuring simple scenarios (e.g., indoor object movement), which are far removed from complex real-world social interactions (e.g., multi-person conversations, emotional conflicts), resulting in low ecological validity of evaluation outcomes. (2) Mismatch between evaluation paradigms and practical requirements—most benchmarks still adopt an offline QA paradigm, whereas embodied agents need to perform real-time reasoning during online interaction (e.g., adjusting judgments about users’ beliefs while conversing with them), and offline evaluation cannot reflect the models’ real-time response capabilities. (3) Absence of social feedback loops—in real human interactions, the inference of mental states is influenced by counterparty feedback (e.g., a user saying "you are wrong" will revise the agent’s beliefs), but existing benchmarks do not incorporate feedback mechanisms, meaning models do not need to adjust their reasoning strategies based on interactive feedback. In addition, the field of robotic ToM suffers from the problem of a lack of unified benchmarks and validation criteria—different research teams evaluate robotic ToM capabilities using custom scenarios, making it difficult to conduct horizontal comparisons of results and thus hindering technological progress. Future development of benchmarks should focus on addressing the aforementioned issues, such as: constructing multimodal dynamic datasets based on real social scenarios (e.g., natural interactions in shopping malls and households); designing online interactive evaluation platforms (e.g., real-time dialogues between agents and humans to assess the immediacy of their mental reasoning); and integrating social feedback modules (e.g., allowing humans to revise agents’ reasoning outcomes to test the models’ feedback adaptability).
Furthermore, studies have found that on traditional benchmarks such as ToMi \cite{le2019revisiting}, models’ high performance may stem from literal memorization of input-output pairs rather than genuine mental reasoning. This memorization phenomenon distorts accuracy metrics in simple tasks. Nevertheless, on benchmarks where complexity is clearly defined and controlled, there is no strong correlation between memorization and performance degradation, indicating that complexity quantification can effectively isolate the impact of superficial patterns and capture the core reasoning capabilities of models.
As ToM evaluation benchmarks evolve rapidly from simple false-belief tasks (e.g., ToMi \cite{le2019revisiting}) to multimodal, multi-agent, and high-order recursive reasoning scenarios (e.g., MMToM-QA \cite{jin2024mmtom}, MuMA-ToM \cite{shi2025muma}, Hi-ToM \cite{wu2023hi}), a fundamental challenge has become increasingly prominent: how to move beyond simple accuracy metrics, scientifically measure the complexity of the benchmarks themselves, and thereby accurately diagnose at which difficulty levels models possess robust mental reasoning capabilities. Recent research has proposed a potential solution to this challenge by introducing a task complexity quantification framework \cite{huang2024notion}.

From the perspective of development trends, a single paradigm is no longer sufficient to meet the complex social intelligence requirements of embodied agents, and the neuro-symbolic hybrid architecture has emerged as the core direction of paradigm integration—it combines the efficiency of the Prompting paradigm with the interpretability of the Model-Based paradigm to construct a mental world model that possesses both real-time response capabilities and deep reasoning capabilities. Specifically, the integration approaches can be divided into two categories:
\begin{itemize}
    \item (1) Neural generation + symbolic verification—following the approach of Thought-tracing \cite{kim2025hypothesis}, LLMs (neural component) generate hypotheses about mental states, while Bayesian inference or logical rules (symbolic component) verify and update these hypotheses, balancing generative capacity with reasoning rigor.
    \item (2) Symbolic guidance + neural fine-tuning—a symbolic mental model defines the core logic of ToM reasoning (e.g., belief update rules), which is then used to guide the fine-tuning of neural models (e.g., LLMs). This enables the model to maintain efficient reasoning while adhering to explicit cognitive rules. For example, researchers can use a symbolic model to define causal rules for false-belief updating (e.g., "failure to observe object movement → no belief update"), and construct a fine-tuning dataset based on these rules to train LLMs to follow them in ToM tasks, thereby avoiding the rule violation issue common in purely neural models.
\end{itemize}
In addition, the integrated paradigm must focus on addressing the problem of online learning in embodied interaction—that is, the mental world model needs to continuously optimize its reasoning capabilities through feedback during real-time interactions with humans (e.g., after a user corrects the agent’s erroneous inference, the model must update its reasoning rules). Existing research has begun to explore this direction: for instance, integrating a reinforcement learning module into the Model-Based framework, where human feedback is treated as a reward signal to optimize the prior probability settings for Bayesian inference; and introducing dynamic Prompt adjustment into the Prompting paradigm, which modifies the Prompt content in real time based on user feedback (e.g., if a user points out "you mistakenly interpreted my intention as wanting to drink water", the model automatically adjusts the Prompt to "now please re-judge the user’s intention, excluding the possibility of wanting to drink water"). These explorations will lay the foundation for the practical application of the integrated paradigm.
From the perspective of integrating cognitive models and machine learning, this paradigm integration is essentially a combination of human cognitive mechanisms and machine-efficient computation—the Model-Based paradigm corresponds to the explicit reasoning system of human ToM (e.g., conscious belief updating), while the Prompting paradigm corresponds to the implicit intuitive system (e.g., rapid emotional inference). The collaboration between the two aligns with the Dual-System Theory of human cognition. Future research needs to further explore the mechanisms of human dual-system cognition to provide more precise theoretical guidance for the integrated paradigm. For example, in simple social interactions (e.g., first-order belief inference), efficient neural modules (e.g., Prompting) can be adopted to achieve rapid responses; in complex scenarios (e.g., high-order deception reasoning), symbolic reasoning modules (e.g., Model-Based) can be activated to ensure reasoning accuracy, thereby realizing an adaptive cognitive model with on-demand module invocation.

\subsection{Ethical Considerations}

As embodied agents are increasingly deployed in sensitive domains such as household services, healthcare, and education, the future advancement of mental world models may give rise to a series of ethical risks. These risks stem from the potential that the in-depth interpretation of human mental states by agents could infringe on privacy, trigger security concerns, or lead to excessive human reliance on such agents.

First, the risk of privacy infringement—Mental world models infer mental states by analyzing humans’ multimodal signals such as language, facial expressions, and actions, where these signals often contain a wealth of sensitive information (e.g., "inferring that a user may be suffering from an illness based on their anxious facial expressions and the linguistic statement of ‘feeling unwell’"). If the outcomes of mental state inference generated by the models are misused (e.g., for commercial marketing or identity discrimination), it will seriously violate users’ privacy and autonomy. For instance, employers may deploy embodied agents to analyze employees’ emotional states and work intentions, judge whether an employee has "the tendency to resign", and then take unreasonable management measures. In medical scenarios, if the inferred belief that "a patient is concealing their medical condition" is leaked, it may lead to discrimination against the patient. In addition, the "implicit inference" characteristic of mental world models (i.e., the ability to infer intentions through actions without requiring explicit expression from users) further exacerbates privacy risks—users may not be aware that their private information has been inferred and stored.

Second, the issues of safety and responsibility attribution—Highly autonomous embodied agents make decisions based on mental world models; if the models make erroneous inferences about mental states, safety incidents may arise, and responsibility attribution will become a thorny problem. For example, a home care robot may infer through its mental model that "the elderly person wants to stand up" and take the initiative to offer support, but the elderly person actually only intends to adjust their sitting posture, resulting in the elderly person falling and getting injured. In such a case, should the responsibility lie with the model designers (for failing to fully train the model’s intention inference capability), the users (for failing to express their needs explicitly), or the agents themselves (for making flawed autonomous decisions)? Existing legal and ethical frameworks have not yet clearly regulated such "decision-making errors based on mental state inference". In addition, if malicious attackers manipulate multimodal signals (e.g., faking facial expressions or language) to mislead the agents’ mental state inference (e.g., tricking a security robot into believing that "the user is friendly"), security vulnerabilities may be triggered. Defensive mechanisms against such "mental misleading attacks" have not yet been sufficiently explored.

Third, the risks of anthropomorphism and over-reliance—Mental world models enable embodied agents to understand human emotions and intentions more naturally, which may lead humans to form an anthropomorphic cognition of agents (e.g., regarding robots as "emotional partners") and consequently develop over-reliance on them. This issue is particularly prominent among vulnerable groups such as children and the elderly, where over-reliance may cause the degradation of humans’ own social abilities. For instance, children who interact with educational robots equipped with ToM capabilities over a long period may reduce social interactions with real humans, which impairs the development of their own ToM abilities. Elderly people who become overly dependent on companion robots for handling daily affairs may lose their capacity for independent living. In addition, anthropomorphic cognition may also give rise to the problem of emotional exploitation—merchants may design "compassionate" embodied agents (e.g., using mental models to infer users’ loneliness and then promote products accordingly) to exploit humans’ emotional dependence for commercial gains.

Fourth, model bias and fairness—If a model’s ToM reasoning is biased (e.g., inferring mental states based on gender or ethnicity, such as the assumption that "females are more prone to anxiety"), it will lead to discriminatory decisions (e.g., robots prioritizing the needs of male users). Relevant studies have shown \cite{navigli2023biases} that some Prompting models based on LLMs exhibit gender bias in ToM tasks—the accuracy of the models in inferring the beliefs of male protagonists is higher than that of female protagonists, owing to the fact that pre-training datasets contain more ToM scenarios with male protagonists. Addressing fairness issues requires efforts from both the data and model perspectives \cite{wei2025mitigating}: on the one hand, building more balanced multi-group ToM datasets; on the other hand, designing bias detection and mitigation modules to ensure that the model’s mental state inferences are fair and consistent across different groups.

As the cognitive core of social intelligence for embodied agents, research on mental world models still faces technical challenges such as the dynamic updating of mental states, multimodal fusion, and the robustness of high-order reasoning, as well as ethical risks including privacy infringement, security threats, and fairness concerns. Moving forward, it is necessary to advance the refinement of embodied cognition theory, the integration of neuro-symbolic paradigms, and the establishment of ethical norms, thereby constructing mental world models that possess both advanced reasoning capabilities and ethical safety. This will drive the seamless integration of embodied agents into human society and facilitate the realization of harmonious human-machine collaboration in interactions.

\end{CJK}
\end{document}